\documentclass[letterpaper]{article} 
\usepackage[preprint]{aaai2027}
\usepackage[hyphens]{url}  
\usepackage{graphicx} 
\usepackage{natbib}  
\usepackage{caption} 
\usepackage{algorithm}
\usepackage{algorithmic}

\usepackage{newfloat}
\usepackage{listings}
\DeclareCaptionStyle{ruled}{labelfont=normalfont,labelsep=colon,strut=off} 
\floatstyle{ruled}
\newfloat{listing}{tb}{lst}{}
\floatname{listing}{Listing}

\usepackage{booktabs}
\definecolor{modelrowA}{RGB}{226,233,242} 
\definecolor{modelrowB}{RGB}{231,226,245} 
\definecolor{modelgray}{RGB}{240,240,240} 

\newcommand{\std}[1]{\textcolor{gray}{\small \textpm #1}}
\usepackage[table]{xcolor}
\usepackage{arydshln}
\usepackage{multirow}
\usepackage{adjustbox}
\usepackage{enumitem}
\usepackage{amsmath}
\usepackage{amssymb}
\usepackage{xurl}
\usepackage{seqsplit}
\usepackage{makecell}
\usepackage[most]{tcolorbox}

\definecolor{rose}{HTML}{cc0200}
\newcommand{\drop}[1]{\textcolor{rose}{$\downarrow$#1}}
\newcommand{\increase}[1]{\textcolor{green!50!black}{\ensuremath{\uparrow}#1}}
\newcommand{\same}[1]{\textcolor{gray}{\ensuremath{#1}}}

\newtcolorbox{prompt}[1]{
    enhanced,
    breakable,
    colback=blue!3,
    colframe=black!70,
    boxrule=0.5pt,
    arc=2mm,
    left=10pt,
    right=10pt,
    boxsep=5pt,
    fonttitle=\bfseries,
    title=#1
}

\newtcolorbox{greenprompt}[1]{
    enhanced,
    breakable,
    colback=green!4,
    colframe=green!30!black,
    boxrule=0.5pt,
    arc=2mm,
    left=10pt,
    right=10pt,
    boxsep=5pt,
    fonttitle=\bfseries,
    title=#1
}
\def \OURS{SciToolAgent-Evo}
\def \OURDATA{OpenSciToolBench}

\title{\OURS{}: An Ontology-Aware Self-Evolving Agent for Open-World Scientific Tool Acquisition}
\author{
    Yuqi Tang\textsuperscript{\rm 1,2},
    Chenyi Zhou\textsuperscript{\rm 1},
    Libin Wang\textsuperscript{\rm 3},
    Keyan Ding\textsuperscript{\rm 2},
    Qiang Zhang\textsuperscript{\rm 2,3}\corresponding,
    Huajun Chen\textsuperscript{\rm 1,2}\corresponding
}

\affiliations{
    \textsuperscript{\rm 1}College of Computer Science and Technology, Zhejiang University \\
    \textsuperscript{\rm 2}ZJU-Hangzhou Global Scientific and Technological Innovation Center, Zhejiang University \\
    \textsuperscript{\rm 3}ZJU-UIUC Institute, Zhejiang University \\
    yuqi.22@intl.zju.edu.cn
}

\begin{document}

\maketitle

\begin{abstract}
Large language model (LLM) agents have been increasingly adopted in scientific research for organizing and invoking specialized computational tools. 
However, their reliance on predefined tool spaces with static semantics limits their applicability to open-world scientific workflows, where tool requirements, capabilities, and boundaries evolve dynamically.
To this end, we propose \OURS{}, an ontology-aware self-evolving agent for open-world scientific tool acquisition. 
Driven by an evolving memory of skills, experiences, and an ontologized tool graph, it distills generalizable knowledge from contrastive trajectories during accumulation, whereas during inference, it formulates active requests and utilizes a LinUCB-based bandit gate to dynamically balance exploration and exploitation. Once a novel tool is acquired, its scientific ontology is completed online for seamless integration into the known graph.
Moreover, we introduce \OURDATA{}, a benchmark containing 900 realistic tasks across four difficulty levels.
Extensive evaluations show that \OURS{} achieves state-of-the-art performance, validating its robustness and generalization.
\end{abstract}


\section{Introduction}

Large language model (LLM) agents have demonstrated remarkable intelligence in complex interactive tasks ~\cite{min2023recent,vemprala2024chatgpt,liu2024agentbench,tang2025learning}, driving their growing adoption in scientific research to organize and invoke specialized computational tools~\cite{wang2024survey,wu2024chateda,chaudhari2026modular,ghareeb2026multi}. However, real scientific workflows rarely begin with a complete and perfectly specified toolbox~\cite{mekala2024toolverifier}. For example, a scientific agent may be asked to identify a molecule satisfying multiple constraints, but the required workflow may involve a descriptor calculator or conversion tool absent from its known tool space. In such cases, closed-world agents are confined to existing tools and often fail when the required capabilities are missing.

This observation suggests a shift from closed-world tool selection to open-world scientific tool acquisition. Advanced scientific tool-use frameworks, such as ChemCrow~\cite{m2024augmenting}, GeneGPT~\cite{jin2024genegpt}, ChatMOF~\cite{kang2024chatmof}, and SciToolAgent~\cite{ding2025scitoolagent} can retrieve, select, and invoke tools from large-scale libraries~\cite{tang2023toolalpaca}. However, their reliance on predefined tool spaces with static semantics hinders capability-gap recognition, leading them to passively over-exploit the known-tool space even when it is insufficient. 
Moreover, without a self-evolving consolidation mechanism, they struggle to reliably ground and reuse newly acquired tool knowledge in future tasks.

Meanwhile, existing scientific tool-use benchmarks~\cite{ma2024sciagent,chen2025scienceagentbench,ding2025scitoolagent} typically involve short tool chains, with task descriptions that contain explicit references to tool names.
Such designs make them insufficient for evaluating the core capabilities required for open-world scientific tool acquisition: recognizing capability gaps, balancing known-tool exploitation with unknown-tool exploration, and continually consolidating tool knowledge.


To address these challenges, we propose \OURS{}, an ontology-aware self-evolving agent for open-world scientific tool acquisition. 
Specifically, \OURS{} maintains an evolving memory consisting of a skill library, an experience pool, and an ontologized tool graph for continual knowledge consolidation and reuse. During accumulation, the agent distills compact tool-use knowledge from contrastive trajectories. At inference time, it formulates active requests and employs a LinUCB-based bandit gate to balance exploration and exploitation. Once an unknown tool is acquired, its ontology is completed online and integrated into the known graph.
Additionally, we construct \OURDATA{}, a hierarchical evaluation benchmark for open-world scientific tool use.
Extensive experiments across scientific tool-use benchmarks demonstrate that \OURS{} exhibits strong robustness and generalization. Our main contributions are as follows:

\begin{itemize}
    \item We propose \textbf{\OURS{}}, an ontology-aware self-evolving agent for open-world scientific tool acquisition. It integrates an evolving memory, active tool requests, and an exploration--exploitation bandit gate to enable autonomous adaptation in dynamic scientific environments.
    
    \item We develop \textbf{\OURDATA{}}, a hierarchical benchmark for open-world scientific tool use, containing 900 realistic tasks across four difficulty levels.
    
    \item We conduct extensive evaluations to demonstrate the superiority of \OURS{} over competitive baselines. Detailed analyses further validate its core mechanisms, including evolving-memory consolidation, active tool requests, and the bandit gate.
\end{itemize}


\section{Related Work}

\subsection{LLM Agents} 

LLM-based agents have shown strong abilities in complex tool-use tasks~\cite{shen2023hugginggpt,paranjape2023art,wang2025otc,qian2026toolrl}. 
Frameworks such as ReAct~\cite{yao2023react}, Reflexion~\cite{shinn2023reflexion}, Toolformer~\cite{schick2023toolformer}, and ToolLLM~\cite{qin2024toolllm} improve tool planning and API matching through in-context learning, feedback-driven refinement, or targeted fine-tuning. 
Recent work further extends agents to scientific domains, with systems such as SciToolAgent~\cite{ding2025scitoolagent} and SciAgent~\cite{ma2024sciagent} automating specialized computation and disciplinary reasoning.
However, these methods largely act as passive selectors within predefined scientific tool spaces, leaving capability-gap recognition, unknown-tool exploration, and continual knowledge consolidation unresolved.
We aim to fill this gap by introducing \OURS{}, an ontology-aware self-evolving agent for open-world scientific tool acquisition.

\subsection{Tool-Use Benchmarks}
In recent years, the development of comprehensive tool-use benchmarks has gained increasing attention ~\cite{mekala2024toolverifier,lu2025toolsandbox,chen2025scienceagentbench}.
Early benchmarks, such as APIBank~\cite{li2023api} and RESTBench~\cite{song2023restgpt}, mainly evaluate static tool selection and invocation under explicit specifications. 
Subsequent benchmarks improve realism by introducing larger tool spaces and unseen-tool generalization; representative examples include Gorilla~\cite{patil2024gorilla}, ToolGen~\cite{wang2025toolgen}, and AppWorld~\cite{trivedi2024appworld}.  
In the scientific domain, SciToolEval~\cite{ding2025scitoolagent} evaluates agents' multi-step API planning and execution capabilities, while SciToolBench~\cite{ma2024sciagent} assesses agents' ability to retrieve and reason with composable scientific functions. 
Despite their effectiveness, these benchmarks typically involve short tool chains and task descriptions with explicit tool-name cues, limiting their ability to evaluate the core capabilities required for open-world scientific tool acquisition.
To this end, we construct \OURDATA, a hierarchical benchmark for open-world scientific tool use.


\section{OpenSciToolBench}
This section presents the dataset construction process of \OURDATA, including its benchmark taxonomy, question generation process, and quality control protocol.

\begin{figure}[t]
    \centering
    \includegraphics[width=\linewidth]{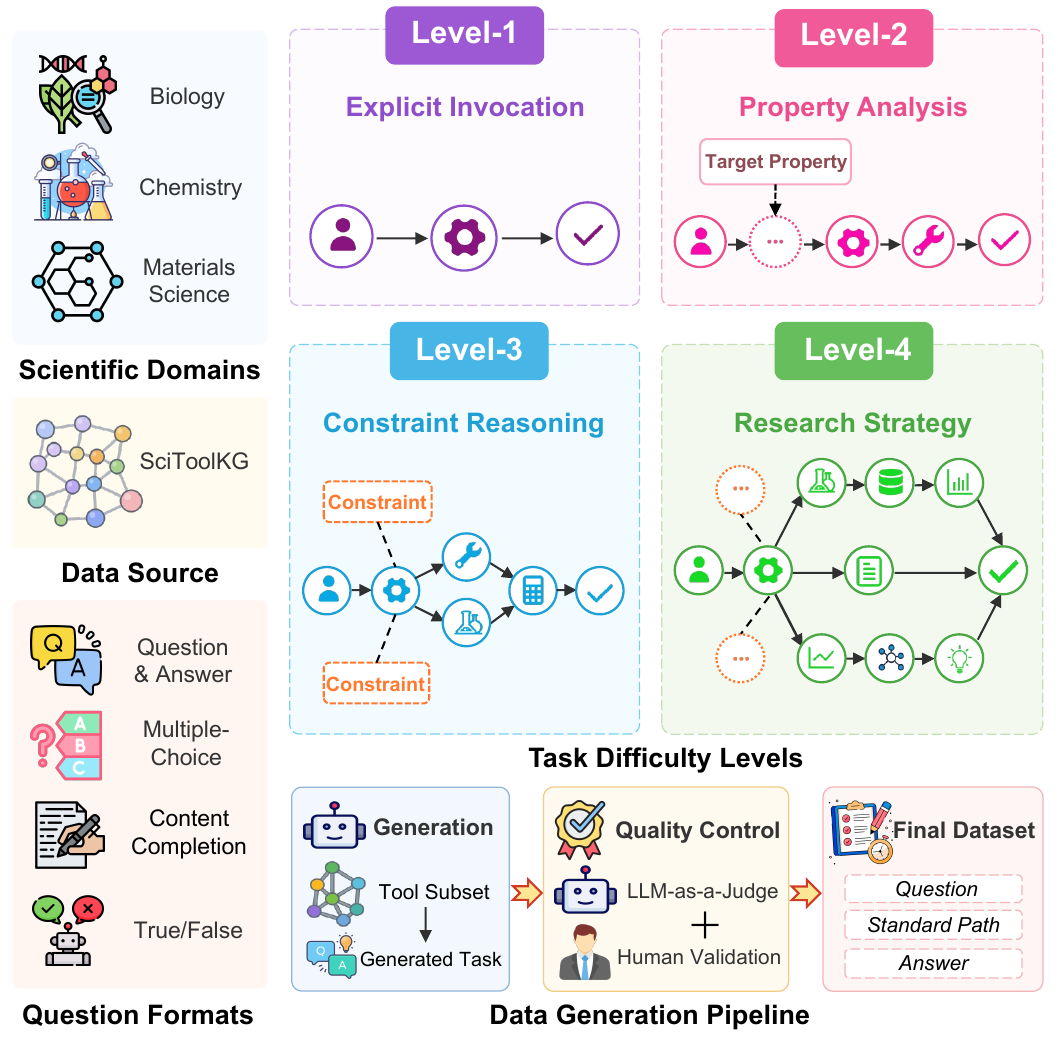}
    \caption{Overview of \OURDATA{}. The benchmark contains 900 scientific tool-use questions constructed from SciToolKG, spanning three scientific domains, four question formats, and four task difficulty levels. The lower panel illustrates the data generation pipeline, including question generation and quality control to construct the final dataset.}
    \label{fig:openscitoolbench_overview}
\end{figure}

\subsection{Benchmark Taxonomy}
Inspired by the Cognitive Rigor Matrix~\cite{hess2009cognitive}, we formulate four difficulty levels to characterize agents' capability boundaries in open-world scientific tool-use scenarios:

\begin{itemize}[leftmargin=*, itemsep=0.5em, topsep=0.5em, parsep=0pt]
    \item \textbf{Level-1 (1--2 tools)}: Assessing explicit tool invocation under directly specified tool requirements.
    \item \textbf{Level-2 (2--4 tools)}: Evaluating property analysis through direct retrieval, computation, or comparison.
    \item \textbf{Level-3 (3--6 tools)}: Evaluating constraint-aware reasoning that integrates multi-dimensional properties under multiple scientific constraints.
    \item \textbf{Level-4 (5--10 tools)}: Challenging open-ended research strategy formulation under realistic research goals.
\end{itemize}

These four levels form the foundation of \OURDATA\ and provide a systematic framework for evaluating open-world scientific tool use.

\subsection{Dataset Generation}
Building on SciToolKG~\cite{ding2025scitoolagent}, which encompasses approximately 500 scientific tools across biology, chemistry, and materials science, we first randomly sample small tool subsets, ensuring that tools within each subset belong to the same domain and are connected through input-output compatibility.
We organize their node information (e.g., descriptions, input/output types, sources) into task specifications. We then craft level-specific prompts and use GPT-5.1~\cite{openai2025gpt5systemcard} to select reasonable tool paths from each subset and generate corresponding questions, ensuring that the generated tasks are scientifically meaningful and aligned with the required competencies.

Detailed prompts are provided in Appendix~\ref{app:generation_prompts}. The resulting questions span four formats: question \& answer, multiple choice, content completion, and true/false, enabling a comprehensive evaluation of open-world scientific tool use.



\subsection{Quality Control}
To ensure scientific rigor and tool-chain coherence, we adopt a two-stage verification process:

\begin{itemize}[left=0pt]
    \item \textit{LLM-as-a-Judge.} We used advanced LLM (e.g., Gemini-3.1-Pro) as the judge to automatically evaluate question clarity, task--tool alignment, and the validity of input--output dependencies within each reference tool chain. Only instances receiving a ``Yes'' judgment were retained.
    \item \textit{Human Validation.} Student reviewers and domain experts then assessed the filtered samples using four predefined criteria: (1) Level Alignment, (2) Semantic Clarity, (3) Tool-Chain Coherence, and (4) Answerability. Only samples approved by a majority of the reviewers were accepted.
\end{itemize}
Based on the process described above, we construct the final \OURDATA, which contains 900 questions across four difficulty levels, multiple scientific domains, and four question formats. More details about the validation protocol and dataset statistics can be found in Appendix~\ref{app:benchmark_validation}.

\section{Methods}

\begin{figure*}[t]
    \centering
    \includegraphics[width=1.0\textwidth]{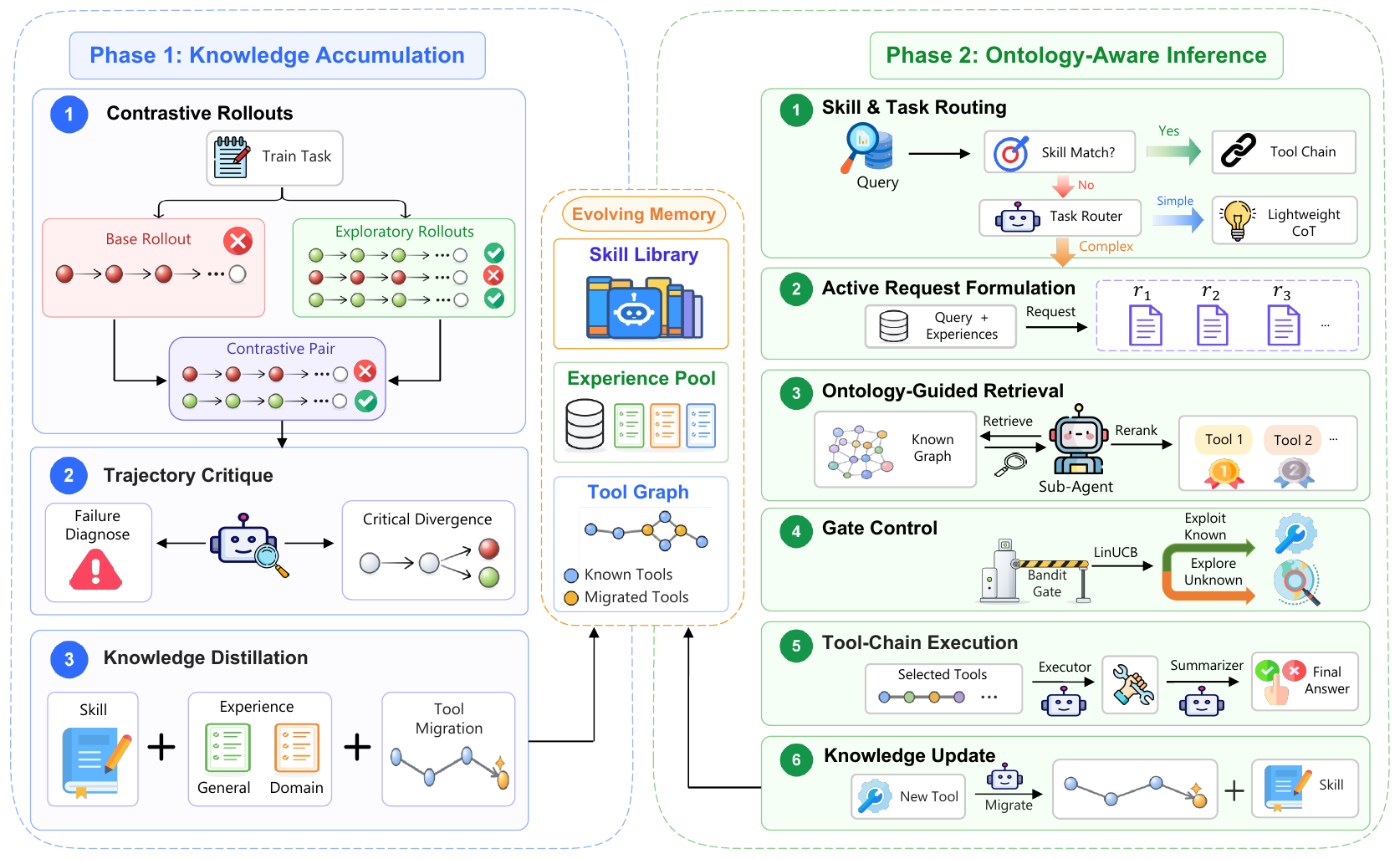}
    \caption{
    Overview of the \OURS{} framework.
    \textbf{Phase I} (left) performs knowledge accumulation over training tasks through contrastive rollout generation, trajectory critique, and knowledge distillation.
    \textbf{Phase II} (right) performs ontology-aware inference through skill and task routing, active request formulation, ontology-guided retrieval, LinUCB-based gate control, tool-chain execution, and knowledge update.
    Together, the two phases continuously enrich a shared evolving memory, which consists of a skill library, an experience pool, and a tool graph.
    }
    \label{fig:scitoolagent_evo_overview}
\end{figure*}

In this section, we introduce \OURS{}, an ontology-aware self-evolving agent for open-world scientific tool acquisition. As illustrated in Figure~\ref{fig:scitoolagent_evo_overview}, \OURS{} consists of two phases: (1) \textit{Knowledge Accumulation}. Given a set of training tasks $\mathcal{D}_{train}$, the agent performs deterministic baseline sampling and temperature-based exploratory sampling, and distills reusable knowledge from contrastive trajectories. (2) \textit{Ontology-Aware Inference}. For a test task $q_{\mathrm{test}}$, the agent retrieves memory, formulates active tool requests, and uses a LinUCB-based bandit gate to acquire tools for executable tool-chain construction.

\subsection{Problem Formulation}

Given a scientific tool-use task $q$, the agent aims to construct an executable tool chain 
\begin{equation}\label{eq:1}
\tau = (t_1, t_2, \ldots, t_m).
\end{equation}
It then derives the final answer via sequential execution, denoted as 
\begin{equation}\label{eq:2}
\hat{y}=\mathrm{Exec}(q,\tau). 
\end{equation}
Let $\mathcal{T}$ denote the complete tool space, which is partitioned into a known tool set $\mathcal{T}_K$ and an unknown tool set $\mathcal{T}_U$. Tools within $\mathcal{T}_K$ are organized as a known tool graph $G_K=(V_K,E_K)$, whereas unknown tools remain outside this graph structure:
\begin{equation}\label{eq:3}
\mathcal{T}=\mathcal{T}_K\cup\mathcal{T}_U,\quad
\mathcal{T}_K\cap\mathcal{T}_U=\emptyset.
\end{equation}
During each planning step, the agent dynamically decides whether to exploit known tools from $G_K$ or explore unknown tools within $\mathcal{T}_U$. To formalize this open-world scientific tool acquisition process, we introduce the following four core components.

\vspace{-0.25em}
\noindent \paragraph{\textbf{Definition 1 (Ontology).}}
An ontology $o \in \mathcal{O}$ is a structured representation that characterizes the semantics, invocation constraints, and scientific applicability of a tool in the known tool graph. Formally, 
$o_t=(b_t,h_t)$, where $b_t$ denotes basic attributes, including functionality, input and output types, and category; 
$h_t$ denotes advanced scientific attributes, including applicable objects and scientific scenarios.
To simulate realistic open-world settings, each unknown tool $u \in \mathcal{T}_U$ is initially under-specified, with only a minimal set of basic attributes exposed.

\vspace{-0.25em}
\noindent \paragraph{\textbf{Definition 2 (Skill).}}
A skill $k_i \in \mathcal{K}$ is a task-level tool-chain guidance unit that provides a generalizable workflow for a specific class of tasks. Each skill is defined as $k_i=(N_i,Z_i,E_i)$, where $N_i$ denotes the skill name, $Z_i$ is a concise description of the skill, and $E_i=(t_1,t_2,\ldots,t_m)$ represents reusable tool chains \cite{jiang2026xskill}.

\vspace{-0.25em}
\noindent \paragraph{\textbf{Definition 3 (Experience).}}
An experience $e \in \mathcal{E}$ is a non-parametric prompt that provides fine-grained guidance for tool selection. Each experience is represented as $e=(d_e,s_e,a_e,h_e)$, where $d_e$ denotes the associated domain, $s_e$ specifies the applicable scientific situation, $a_e$ gives the corresponding tool-selection guidance, and $h_e \in \mathbb{R}^{d_h}$ denotes its retrieval embedding.

\vspace{-0.25em}
\noindent \paragraph{\textbf{Definition 4 (Bandit Gate).}}
A bandit gate is a LinUCB-based~\cite{li2010contextual} decision module that balances the inherent open-world exploration--exploitation trade-off. For each tool request $r_j$, the gate constructs an ontology-guided, four-dimensional context vector $\mathbf{x}_j=[f_j,m_j,s_j,p_j]^\top$, where $f_j$ denotes the top-1 known-tool feasibility, $m_j$ denotes the confidence margin between the top-2 candidates, $s_j$ denotes the ontology alignment score, and $p_j$ denotes the historical success prior of unknown-tool exploration.

With these definitions, we model each open-world scientific tool acquisition task as a Partially Observable Markov Decision Process (POMDP).
Let $q$ denote a task instance. The agent is equipped with a known tool graph $G_K=(V_K,E_K)$, which contains structured tool attributes and tool-level relations. 
At each time step $t$, the agent receives an observation $o_t$ of tools retrieved from the graph, and generates an action $a_t$ based on the current state $s_t$. The action corresponds to either known-tool exploitation or unknown-tool exploration. This process yields a trajectory $\xi = \big[(s_0,a_0,o_0), \ldots, (s_T,a_T,o_T)\big]$,
which finally concludes with execution and summarization to produce the final answer $\hat{y}$.
The objective is to construct an evolving memory $\mathcal{M}=(G_K,\mathcal{K},\mathcal{E})$ that, when combined with the agent, executor, and summarizer, maximizes the probability of generating the correct answer: $\max_{\mathcal{M}} P(\hat{y}=y^* \mid q,\mathcal{M})$.

\subsection{Knowledge Accumulation}
In long-horizon tool-use tasks, a single failed trajectory is too verbose and noisy to determine whether an error arises from a policy flaw or stochastic execution noise. Therefore, without updating the backbone LLM parameters, \OURS{} distills tool-use knowledge from contrastive rollouts.

\noindent \paragraph{\textbf{Contrastive Rollout Generation.}}
For each training task $q_i \in \mathcal{D}_{train}$, the agent first generates the trajectory $\xi_i^{base}$ at temperature $T_{\mathrm{base}}$.
If $\xi_i^{base}$ succeeds, the system extracts a skill $k_i$ abstracted from the trajectory and stores it in $\mathcal{K}$.
Otherwise, the agent rolls out $N-1$ additional trajectories 
$\{\xi_i^{}\}_{i=1}^{N-1}$ 
at different temperatures 
$T_{\mathrm{exp}}=\{\gamma_n\}_{n=1}^{N-1}$.
If any exploratory trajectory succeeds, the system samples one successful trajectory $\xi_i^{+}$ and constructs the contrastive pair $(\xi_i^{base},\xi_i^{+})$.

\noindent \paragraph{\textbf{Rollout Critique and Knowledge Distillation.}}
Given the contrastive pair $(\xi_i^{base},\xi_i^{+})$, the agent first identifies the critical divergence and produces a diagnostic text $C_i$ to explain the corresponding failure reason.
Based on $C_i$, the agent further distills two types of tool-selection experiences, namely domain-specific experiences $e_i^{\mathrm{dom}}$ and general experiences $e_i^{\mathrm{gen}}$.
Before adding them to the experience pool, the system checks whether any existing entry exhibits a cosine similarity above $\theta_{\mathrm{sim}}$ with the candidate experience $e$. If so, the candidate experience is discarded; otherwise, $e$ is added directly. Detailed prompts for skill extraction and experience distillation are provided in Appendix~\ref{app:workflow_prompts}.

\begin{table*}[!ht]
\centering
\renewcommand{\arraystretch}{1.2}
\setlength{\tabcolsep}{1.1mm}
\resizebox{1.0\textwidth}{!}{
\begin{tabular}{lcccccccccc}
\toprule
\multirow{2}{*}{\bf \textsc{Method}} & \multicolumn{2}{c}{\bf \textsc{Level-1}} & \multicolumn{2}{c}{\bf \textsc{Level-2}} & \multicolumn{2}{c}{\bf \textsc{Level-3}} & \multicolumn{2}{c}{\bf \textsc{Level-4}} & \multicolumn{2}{c}{\bf \textsc{Overall}} \\
\cmidrule(lr){2-3} \cmidrule(lr){4-5} \cmidrule(lr){6-7} \cmidrule(lr){8-9} \cmidrule(lr){10-11}
& \bf \textsc{Tool} & \bf \textsc{Ans.} & \bf \textsc{Tool} & \bf \textsc{Ans.} & \bf \textsc{Tool} & \bf \textsc{Ans.} & \bf \textsc{Tool} & \bf \textsc{Ans.} & \bf \textsc{Tool} & \bf \textsc{Ans.} \\
\midrule

\rowcolor{modelrowA} \multicolumn{11}{c}{\raisebox{-0.15ex}{\includegraphics[height=0.9em]{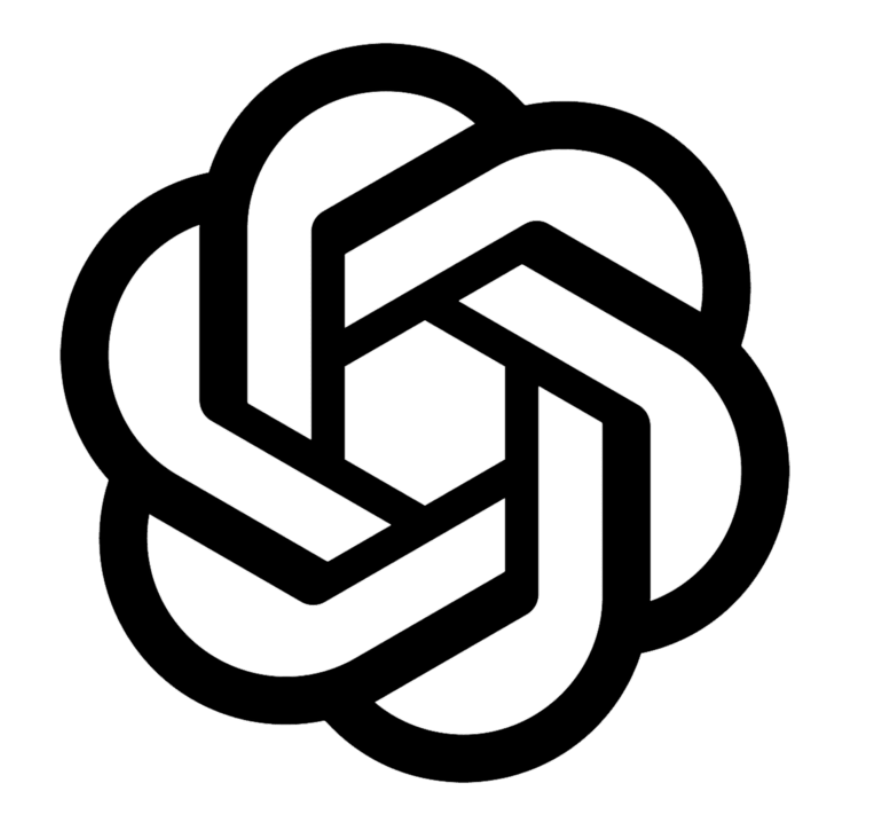}} \textit{GPT-5.4-mini}} \\
\addlinespace[2pt]
Direct       & 52.66\std{1.69} & 51.43\std{1.34} & 46.31\std{1.49} & 43.68\std{0.74} & 24.71\std{1.66} & 21.18\std{3.12} & 23.03\std{2.79} & 21.05\std{2.27} & 38.11\std{0.86} & 35.87\std{0.42} \\
CoT          & 75.24\std{0.00} & 73.33\std{0.95} & \underline{66.32}\std{2.98} & \underline{64.56}\std{1.61} & 38.83\std{1.66} & \underline{36.08}\std{2.04} & \underline{28.29}\std{4.65} & \underline{25.43}\std{3.72} & 54.44\std{0.59} & \underline{52.17}\std{0.89} \\
RAG          & 64.76\std{0.95} & 62.54\std{1.46} & 57.89\std{3.16} & 55.44\std{3.22} & 31.76\std{2.04} & 28.23\std{3.11} & 19.74\std{2.63} & 18.42\std{2.63} & 45.70\std{0.48} & 43.31\std{1.37} \\
ReAct        & 74.06\std{0.67} & 71.31\std{0.67} & 58.95\std{2.98} & 54.21\std{3.72} & 30.58\std{1.17} & 25.88\std{0.83} & 23.68\std{3.72} & 19.74\std{1.32} & 49.45\std{1.56} & 45.26\std{0.64} \\
Reflexion    & \underline{78.30}\std{1.34} & \underline{76.19}\std{1.34} & 64.39\std{4.87} & 60.34\std{3.38} & \underline{40.10}\std{3.02} & 35.29\std{2.36} & 26.32\std{3.72} & 20.62\std{2.01} & \underline{54.70}\std{2.16} & 50.69\std{1.39} \\
SciToolAgent & 53.33\std{2.69} & 51.91\std{1.98} & 48.42\std{1.16} & 47.37\std{3.01} & 33.94\std{2.77} & 30.58\std{2.77} & 17.11\std{1.86} & 14.91\std{3.04} & 39.85\std{0.70} & 37.90\std{1.73} \\
\addlinespace[2pt]
\cdashline{1-11}
\textbf{\OURS} & \textbf{85.71}\std{0.00} & \textbf{83.49}\std{0.55} & \textbf{76.84}\std{3.01} & \textbf{75.26}\std{2.23} & \textbf{58.82}\std{1.65} & \textbf{56.32}\std{1.17} & \textbf{36.84}\std{4.13} & \textbf{32.89}\std{3.48} & \textbf{66.76}\std{0.74} & \textbf{64.28}\std{0.85} \\
\addlinespace[2pt]
\midrule

\rowcolor{modelrowB} \multicolumn{11}{c}{\raisebox{-0.15ex}{\includegraphics[height=0.9em]{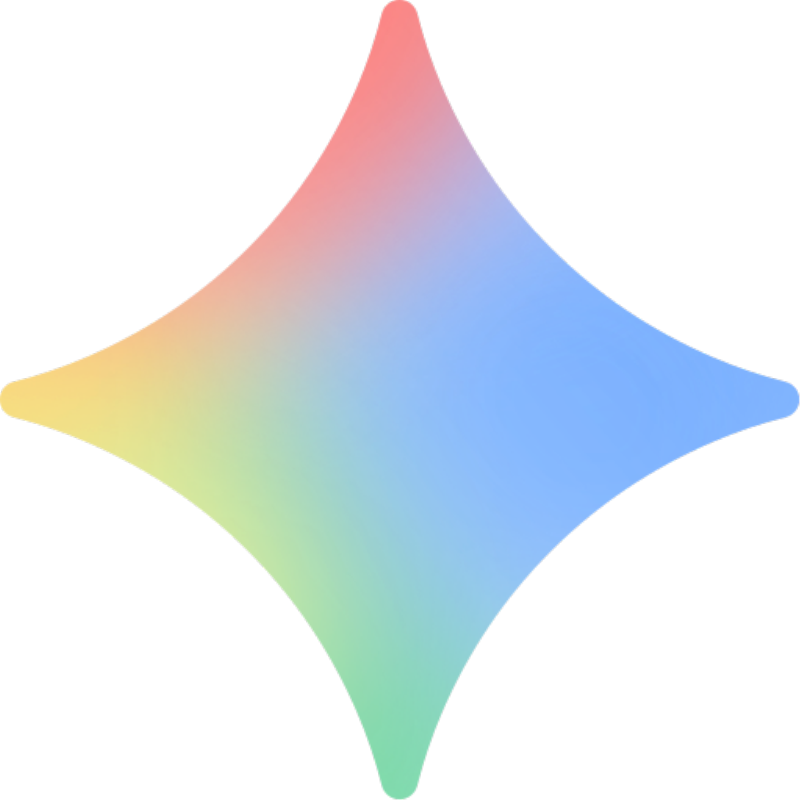}} \textit{Gemini-3-flash}} \\
\addlinespace[2pt]
Direct       & 43.34\std{0.67} & 40.64\std{0.67} & 28.42\std{1.48} & 27.37\std{1.05} & 17.06\std{0.83} & 16.08\std{0.68} & 6.58\std{0.00} & 6.14\std{0.76} & 25.48\std{0.39} & 24.10\std{1.05} \\
CoT          & 57.14\std{1.34} & 56.51\std{1.61} & 43.16\std{3.08} & 42.45\std{3.08} & 30.59\std{1.66} & 29.80\std{0.68} & 26.32\std{1.86} & 24.12\std{1.86} & 40.72\std{0.55} & 39.70\std{0.64} \\
RAG          & 65.71\std{0.96} & 62.86\std{0.00} & 47.37\std{1.05} & 45.26\std{2.11} & 26.27\std{2.96} & 23.92\std{1.80} & 17.11\std{0.00} & 15.35\std{0.76} & 41.37\std{0.64} & 39.06\std{0.83} \\
ReAct        & 75.24\std{1.91} & 72.07\std{1.46} & 58.78\std{2.73} & 53.68\std{2.23} & 41.51\std{3.69} & 36.47\std{3.12} & 27.63\std{0.76} & 23.68\std{0.00} & 52.94\std{2.27} & 48.66\std{1.12} \\
Reflexion    & \textbf{84.43}\std{0.95} & \underline{80.31}\std{2.40} & \underline{67.17}\std{2.87} & \underline{63.51}\std{4.26} & \underline{54.44}\std{3.60} & \underline{48.24}\std{1.36} & \underline{34.21}\std{3.95} & \underline{29.82}\std{1.52} & \underline{62.25}\std{1.37} & \underline{57.71}\std{1.21} \\
SciToolAgent & 63.81\std{2.43} & 62.54\std{1.46} & 53.58\std{3.28} & 53.16\std{2.23} & 43.53\std{0.47} & 41.76\std{2.49} & 22.76\std{4.93} & 21.05\std{3.95} & 47.65\std{2.45} & 46.44\std{2.10} \\
\addlinespace[2pt]
\cdashline{1-11}
\textbf{\OURS} & \underline{83.81}\std{0.95} & \textbf{82.22}\std{1.32} & \textbf{82.11}\std{2.10} & \textbf{79.65}\std{2.10} & \textbf{68.24}\std{5.43} & \textbf{65.09}\std{3.78} & \textbf{35.53}\std{2.79} & \textbf{33.55}\std{0.93} & \textbf{69.53}\std{2.77} & \textbf{67.26}\std{0.96} \\
\bottomrule
\end{tabular}
}
\caption{Evaluation results on \textbf{OpenSciToolBench} with \textbf{GPT-5.4-mini} and \textbf{Gemini-3-flash} backbones. Results are averaged over 3 independent runs and reported as mean $\pm$ standard deviation. The best result is \textbf{bolded}, and the second best is \underline{underlined}. \textsc{Tool} and \textsc{Ans.} indicate tool-use and final answer accuracy.}
\label{main_experiment1}
\end{table*}

\subsection{Ontology-Aware Inference}

\noindent \paragraph{\textbf{Skill and Task Routing.}}
Given a test task $q$, the agent first checks if a relevant skill can be retrieved from $\mathcal{K}$. If a matched skill is found, it executes its tool chain and summarizes the final answer. Otherwise, the agent estimates difficulty from expected tool and entity counts. Simple tasks are solved with lightweight CoT planning, while complex tasks are routed to active tool acquisition.

\noindent \paragraph{\textbf{Active Tool Acquisition.}}
The agent first retrieves relevant experiences to augment its context. While general experiences $\mathcal{E}_{\mathrm{gen}}$ provide default guidance, domain-specific experiences are retrieved via semantic similarity:
\begin{equation}\label{eq:4}
\mathcal{E}_q=
\mathcal{E}_{\mathrm{gen}}\cup \mathrm{TopK}(\mathcal{E}_{d_q}, q, K).
\end{equation}
Here, $d_q$ denotes the task domain, and $K$ specifies the number of retrieved experiences.

Rather than passively selecting tools based on the raw query, the agent actively formulates tool requests via ontology-guided reasoning. Given $q$ and $\mathcal{E}_q$, it generates a set of active tool requests $R_q=\{r_1,r_2,\ldots,r_l\}$, where each $r_j$ is a concise requirement description for one required tool. We justify this design in Appendix~\ref{app:active_request}.

For each request $r_j$, \OURS{} identifies candidate tools within the known tool graph $G_K$. Let $t\in V_K$ denote a tool node and $S(r_j,t)$ represent its semantic similarity to the request. The initial retrieval set is formulated as:
\begin{equation}\label{eq:5}
C_j^{0}=\mathrm{TopK}\{t\mid S(r_j,t),\forall t\in V_K\}.
\end{equation}
After that, graph exploration and combination reranking refine $C_j^{0}$ into the final candidate set $C_j$.

The sub-agent then utilizes $C_j$ to produce the 4D context vector $\mathbf{x}_j=[f_j,m_j,s_j,p_j]^\top \in \mathbb{R}^4$ for the bandit gate. The gate selects between known-tool exploitation and unknown-tool exploration using a LinUCB policy:
\begin{equation}\label{eq:6}
\resizebox{0.78\linewidth}{!}{$
\begin{aligned}
a_j=\arg\max_{a\in\{\texttt{a1},\texttt{a2}\}}
\left(
\hat{\boldsymbol{\theta}}_a^\top \mathbf{x}_j+
\alpha\sqrt{\mathbf{x}_j^\top A_a^{-1}\mathbf{x}_j}
\right),
\end{aligned}
$}
\end{equation}

\noindent where $\texttt{a1}$ and $\texttt{a2}$ denote known-tool exploitation and unknown-tool exploration, respectively. Here, $\hat{\boldsymbol{\theta}}_a=A_a^{-1}\mathbf{b}_a$ estimates the reward of action $a$, and $\alpha$ controls the exploration strength. 
For exploitation, the agent selects the top-ranked candidate from $C_j$; for exploration, it infers tool utility from sparse attributes and retrieves the most compatible candidate from $\mathcal{T}_U$.
The action is applied iteratively until a complete tool chain $\hat{\tau}$ is constructed. Finally, the Executor runs $\hat{\tau}$ sequentially, and the Summarizer produces $\hat{y}$.

\noindent \paragraph{\textbf{Knowledge Update.}}
Inspired by human learning, successful exploration is consolidated into reusable knowledge. When a rollout successfully utilizes an unknown tool, the agent completes its ontology $o_t$ from the interaction history and migrates it into the known tool graph $G_K$. The same execution signal provides a binary reward $r\in\{0,1\}$ for updating the parameters of the selected action $a$: $A_a \leftarrow A_a + \mathbf{x}_j\mathbf{x}_j^\top, \mathbf{b}_a \leftarrow \mathbf{b}_a + r\mathbf{x}_j$.
In addition, the resulting trajectory is abstracted into a skill $k_{\mathrm{new}}$ and stored in $\mathcal{K}$.
Together, these updates establish a closed self-evolving loop, enabling the agent to adapt continuously in open-world environments.

\begin{table*}[!ht]
\centering
\setlength{\tabcolsep}{3.0mm}
\small
\resizebox{1\textwidth}{!}{
\begin{tabular}{lcccccc} 
\toprule
\multirow{2}{*}{\bf \textsc{Method}} & \multicolumn{2}{c}{\bf \textsc{Level-1}} & \multicolumn{2}{c}{\bf \textsc{Level-2}} & \multicolumn{2}{c}{\bf \textsc{Overall}} \\
\cmidrule(lr){2-3} \cmidrule(lr){4-5} \cmidrule(lr){6-7} 
& \bf \textsc{Tool} & \bf \textsc{Ans.} & \bf \textsc{Tool} & \bf \textsc{Ans.} & \bf \textsc{Tool} & \bf \textsc{Ans.} \\
\midrule

\rowcolor{modelrowA} \multicolumn{7}{c}{\raisebox{-0.15ex}{\includegraphics[height=0.9em]{gpt_logo.pdf}} \textit{GPT-5.4-mini}} \\
\addlinespace[2pt]
Direct       & 53.29\std{0.93} & 51.75\std{1.24} & 46.43\std{1.68} & 45.24\std{1.98} & 49.01\std{0.70} & 47.69\std{0.81} \\
CoT          & 86.84\std{1.86} & 85.53\std{1.32} & \underline{72.62}\std{2.81} & \underline{71.43}\std{2.38} & \underline{77.97}\std{2.45} & \underline{76.73}\std{1.73} \\
RAG          & 76.32\std{1.32} & 74.12\std{0.76} & 48.94\std{1.99} & 48.41\std{2.10} & 59.24\std{1.51} & 58.09\std{1.51}  \\
ReAct        & 83.56\std{3.48} & 82.45\std{2.74} & 69.05\std{3.46} & 68.25\std{3.46} & 74.51\std{2.70} & 73.59\std{2.23} \\
Reflexion    & \underline{87.28}\std{2.28} & \underline{86.39}\std{2.01} & 70.63\std{1.68} & 68.78\std{1.99} & 76.89\std{1.24} & 75.41\std{2.20} \\
SciToolAgent & 62.50\std{2.79} & 61.18\std{2.79} & 46.43\std{1.68} & 45.24\std{1.12} & 52.48\std{2.10} & 51.24\std{1.75} \\
\addlinespace[2pt]
\cdashline{1-7}
\textbf{\OURS} & \textbf{90.13}\std{0.93} & \textbf{88.60}\std{0.76} & \textbf{80.16}\std{3.37} & \textbf{79.37}\std{3.37} & \textbf{83.91}\std{1.75} & \textbf{82.84}\std{1.87} \\
\addlinespace[2pt]
\midrule

\rowcolor{modelrowB} \multicolumn{7}{c}{\raisebox{-0.15ex}{\includegraphics[height=0.9em]{gemini_logo.pdf}} \textit{Gemini-3-flash}} \\
\addlinespace[2pt]
Direct       & 55.26\std{1.86} & 53.95\std{1.86} & 25.00\std{2.80} & 25.00\std{2.80} & 36.37\std{2.45} & 35.89\std{2.45} \\
CoT          & 87.50\std{0.93} & \underline{86.84}\std{0.76} & 51.98\std{2.81} & 50.79\std{2.25} & 65.34\std{2.10} & 64.35\std{1.40} \\
RAG          & 78.07\std{3.31} & 76.75\std{3.31} & 53.44\std{1.21} & 52.38\std{1.59} & 62.71\std{1.59} & 61.55\std{1.87}  \\
ReAct        & 85.53\std{2.78} & 85.09\std{1.52} & \underline{69.84}\std{1.12} & \underline{69.04}\std{0.57} & \underline{75.74}\std{1.75} & \underline{75.08}\std{0.76} \\
Reflexion    & \textbf{88.82}\std{1.32}  & \textbf{88.16}\std{1.42} & 65.48\std{3.96} & 64.68\std{2.71} & 74.26\std{1.51} & 73.51\std{1.03} \\
SciToolAgent & 71.06\std{3.72} & 70.39\std{2.79} & 55.17\std{2.76} & 54.81\std{2.18} & 61.14\std{1.40} & 60.67\std{0.33} \\
\addlinespace[2pt]
\cdashline{1-7}
\textbf{\OURS} & \underline{87.72}\std{0.88} & \underline{86.84}\std{1.87} & \textbf{73.41}\std{1.68} & \textbf{72.62}\std{1.32} & \textbf{78.79}\std{1.40} & \textbf{77.97}\std{1.75} \\
\bottomrule
\end{tabular}
}
\caption{Evaluation results on \textbf{SciToolEval} with \textbf{GPT-5.4-mini} and \textbf{Gemini-3-flash} backbones. Results are averaged over 3 independent runs and reported as mean $\pm$ standard deviation. The best result is \textbf{bolded}, and the second best is \underline{underlined}. \textsc{Tool} and \textsc{Ans.} indicate tool-use and final answer accuracy.}
\label{main_experiment2}
\vspace{-0.5em}
\end{table*}

\section{Experiments}

\subsection{Evaluation Benchmarks and Metrics}

We conduct a comprehensive evaluation of \OURS\ on two scientific tool-use benchmarks:

\begin{itemize}[leftmargin=*, itemsep=0.5em, topsep=0.5em, parsep=0pt]
    \item \textbf{\OURDATA}: Contains 900 scientific tool-use questions to assess capabilities ranging from explicit tool invocation to open-ended research strategy formulation.
    
    \item \textbf{SciToolEval}~\cite{ding2025scitoolagent}: Features 531 questions comprising 152 single-tool and 379 multi-tool tasks, covering molecular property prediction, protein analysis, and materials retrieval.
\end{itemize}
For both benchmarks, we apply \textbf{Tool Accuracy} (measuring tool-chain alignment with the reference) and \textbf{Answer Accuracy} (assessing final-answer correctness) as the final metrics.

\subsection{Baselines}\label{sec:baselines}

We evaluate \OURS\ under two proprietary LLM backbones, GPT-5.4-mini~\cite{openai2025gpt5systemcard} and Gemini-3-flash~\cite{team2023gemini}, and compare it against various representative methods: 
\textbf{Direct} generates the final answer directly without producing intermediate reasoning; \textbf{CoT}~\cite{wei2022chain} produces an intermediate reasoning trajectory before answering; 
\textbf{RAG}~\cite{lewis2020retrieval} retrieves relevant external knowledge to augment answer generation;
\textbf{ReAct}~\cite{yao2023react} and \textbf{Reflexion}~\cite{shinn2023reflexion} serve as interactive tool-use methods, relying on the Thought-Action-Observation loop and trajectory-level self-feedback for planning respectively; and \textbf{SciToolAgent}~\cite{ding2025scitoolagent} leverages a tool knowledge graph for tool retrieval, tool-chain planning, and execution.
More details of these baselines are presented in Appendix~\ref{app:baseline_details}.

\subsection{Implementation Details}

We employ locally deployed Qwen3-Embedding-0.6B and Qwen3-1.7B~\cite{yang2025qwen3} as our retrieval model and reranking model, respectively. For all baselines, we use the same open-world split, with 50\% of the toolset designated as known and the remaining 50\% as unknown. During accumulation, we sample $K=4$ trajectories for each task, including one deterministic baseline trajectory with $T_{\mathrm{base}}=0$ and three exploratory trajectories with $T_{\mathrm{exp}}\in\{0.2,0.5,0.8\}$. At test time, we use a LinUCB-based bandit gate with $\alpha_0=0.8$, decay rate $0.01$, $\alpha_{\min}=0.2$, and a 20-step failure window to scale $\alpha$ within $[0.7,1.3]$. More implementation details are provided in Appendix~\ref{app:experimental_configuration}.

\subsection{Main Results}

Tables~\ref{main_experiment1} and~\ref{main_experiment2} show the results across two benchmarks and two backbone models. Our method demonstrates substantial improvements over all baselines. 
Compared with the runner-up results, \OURS{} improves overall answer accuracy on \OURDATA\ by 12.11 and 9.55 points with GPT-5.4-mini and Gemini-3-flash, respectively. On SciToolEval, the corresponding gains are 6.11 and 2.89 points.
Notably, \OURS{} exceeds SciToolAgent by a striking 17\% to 31\% margin across all benchmarks. 
This gain reflects a key distinction: SciToolAgent remains constrained by a fixed known-tool graph, whereas \OURS{} dynamically balances exploration and exploitation and consolidates acquired knowledge to enable reliable acquisition of unknown tools, yielding stronger open-world adaptability.

From a fine-grained analysis across difficulty levels, we observe that while the performance gap is marginal on simpler tasks (Level-1), \OURS{} demonstrates a significant advantage in solving complex multi-tool tasks. For instance, on \OURDATA\ with GPT-5.4-mini, \OURS{} achieves 56.32\% answer accuracy on Level-3, surpassing Reflexion and ReAct by more than 20 and 30 points. 
Empirically, we observe that in-context learning methods, such as CoT, ReAct, and Reflexion, rely heavily on local step-wise reasoning and lack a stable global strategy for long tool-chain planning. In open tool spaces with vast and initially unobserved candidates, this deficiency inevitably incurs high exploration costs and degrades tool-planning accuracy.
\OURS{} instead maintains an evolving memory, reducing exploration costs and guaranteeing robust adaptation in open-world environments.
\begin{table*}[!t]
\centering
\resizebox{1.0\textwidth}{!}{
\begin{tabular}{lccccc}
\toprule
Method & Level-1 & Level-2 & Level-3 & Level-4 & Overall \\
\midrule
w/o Knowledge Accumulation & 80.00 \drop{5.71} & 57.89 \drop{18.95} & 45.88 \drop{14.93} & 14.47 \drop{22.37} & 52.35 \drop{14.87}  \\
w/o Tool Migration & 78.10 \drop{7.61} & 62.11 \drop{14.73} &  41.17 \drop{19.64} & 13.16 \drop{23.68} & 51.53 \drop{15.69} \\
w/o Inference Migration & 81.90 \drop{3.81} & 72.63 \drop{4.21} & 56.47 \drop{4.34} & 35.53 \drop{1.31} & 63.71 \drop{3.51} \\
w/o Skill  & 83.81 \drop{1.90} & 73.68 \drop{3.16} & 54.12 \drop{6.69} & 35.53 \drop{1.31} & 63.99 \drop{3.23} \\
w/o Experience & 85.71 \drop{0.00} & 75.79 \drop{1.05} & 57.65 \drop{3.16} & 34.21 \drop{2.63} & 65.65 \drop{1.57} \\
\midrule
\rowcolor{modelgray}
\textbf{Ours - Full Pipeline} & 85.71 & 76.84 & 60.81 & 36.84 & 67.22 \\
\bottomrule
\end{tabular}
}
\caption{Ablation study on \textbf{\OURDATA}. All variants use GPT-5.4-mini as the backbone. We remove key components to assess their contributions, and \drop{} denotes the absolute performance drop from the full pipeline.}
\label{tab:ablation_study}
\end{table*}

\begin{figure}[!ht]
    \centering
    \includegraphics[width=0.48\textwidth]{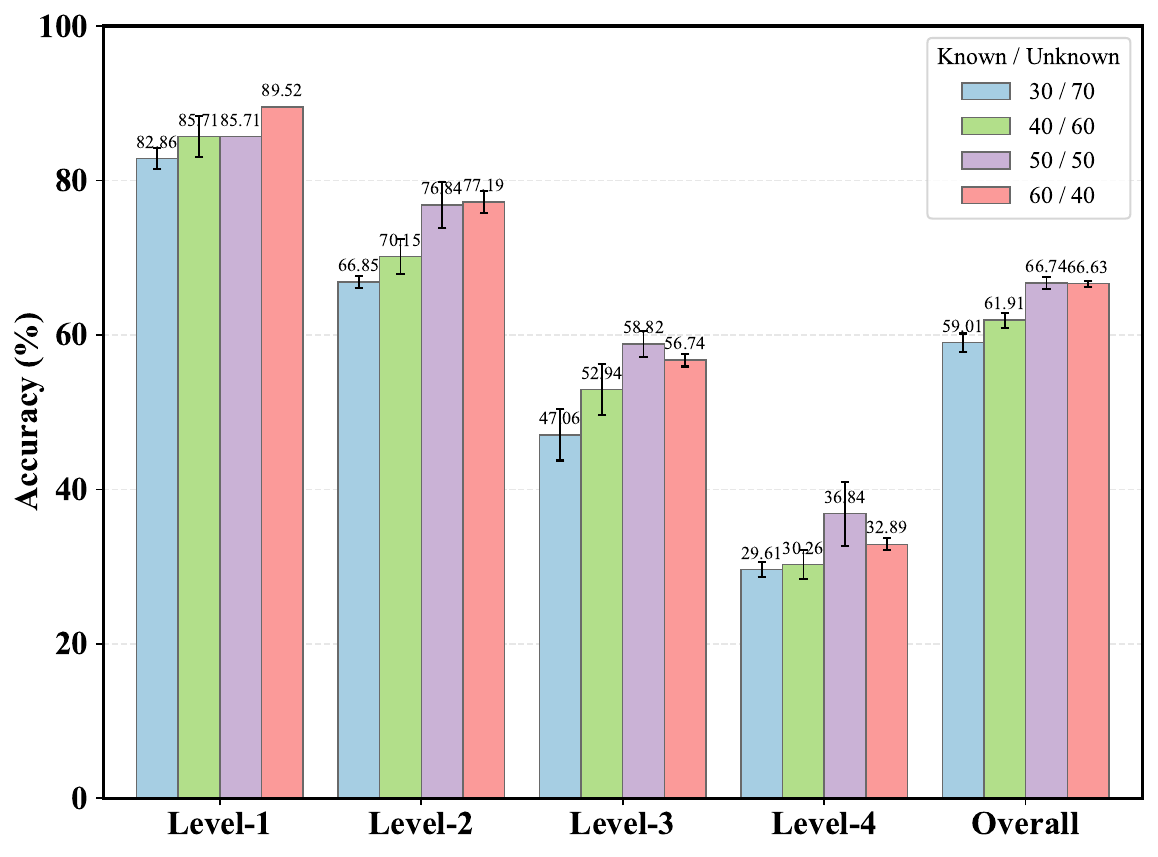}
    \caption{Performance sensitivity to known/unknown tool splits on \textbf{\OURDATA} with GPT-5.4-mini.}
    \label{fig:split_sensitivity}
\end{figure}

\subsection{Ablation Study}\label{sec:Ablation_Study}

\paragraph{Component Ablation.}

We conduct a systematic ablation study on \OURDATA{} with GPT-5.4-mini to analyze the contribution of each component. 
As shown in Table~\ref{tab:ablation_study}, we obtain the following results: 
(1) Removing knowledge accumulation causes a 14.87-point drop, showing its necessity for robust adaptation in open-world scientific tool settings.
(2) Disabling unknown tool migration entirely leads to the largest degradation. Conversely, disabling it only during inference causes a much smaller drop, underscoring the necessity of continual tool migration.
(3) Excluding skills and experiences degrades performance by 3.23 and 1.57 points, respectively, demonstrating that both knowledge forms are critical.
These results indicate that each component addresses a distinct challenge in open-world scientific tool use, 
and their integration enables continual refinement of tool memory and acquisition policy. Further fine-grained ablation results for the inference pipeline are reported in Appendix~\ref{app:fine_grained_ablation}.

\paragraph{Known/Unknown Split Sensitivity.}
To further validate the robustness of our method, we evaluate \OURS{} on \OURDATA{} across four known/unknown tool splits using GPT-5.4-mini. As shown in Figure~\ref{fig:split_sensitivity}, overall accuracy generally improves as the known-tool ratio grows, rising from 59.01\% under the 30/70 split to 66.63\% under the 60/40 split. This suggests that a richer known-tool graph reduces open-world exploration difficulty and stabilizes tool-chain planning. 
Crucially, even when 70\% of tools are initially unknown, \OURS{} still achieves a competitive 59.01\% overall accuracy.
This result demonstrates the framework's robust tool-acquisition capability in challenging open-world settings.

\begin{figure}[!ht]
    \centering
    \begin{minipage}[t]{0.49\linewidth}
        \centering
        \includegraphics[width=\linewidth]{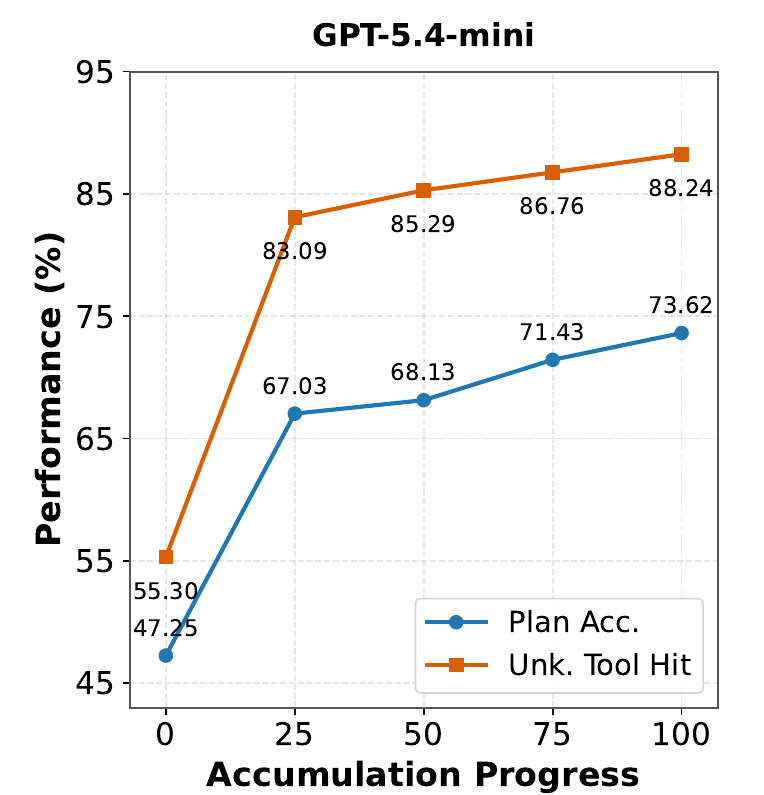}
    \end{minipage}
    \hfill
    \begin{minipage}[t]{0.49\linewidth}
        \centering
        \includegraphics[width=\linewidth]{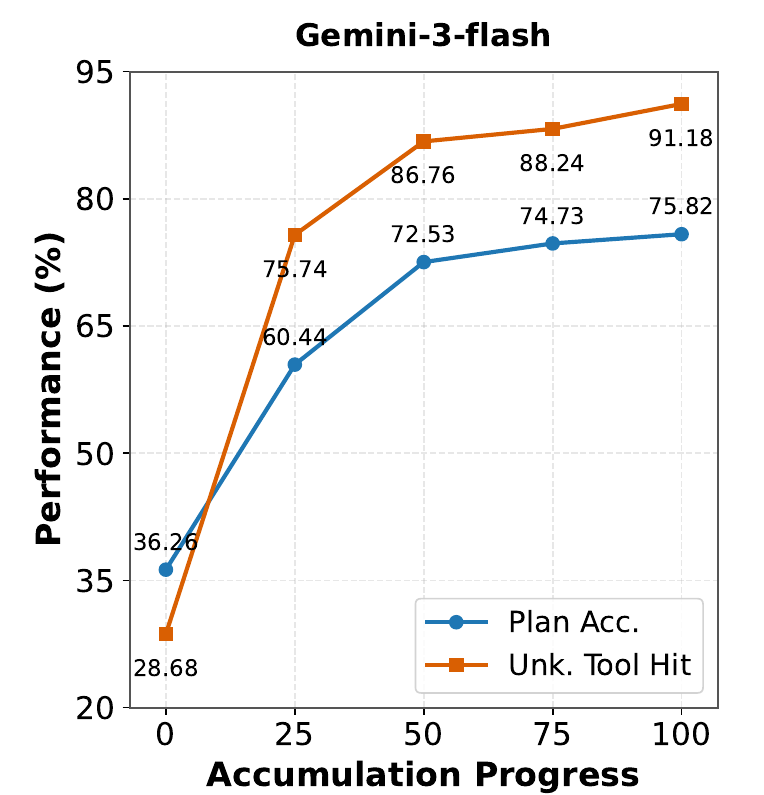}
    \end{minipage}

    \caption{Performance comparison across knowledge accumulation stages on \textbf{\OURDATA} with GPT-5.4-mini and Gemini-3-flash, measured by Plan Acc. (planning accuracy) and Unk. Tool Hit (hit rate on initially unknown tools).}
    \label{fig:self_evolve_dynamics}
\end{figure}

\subsection{Knowledge Accumulation Dynamics}

To investigate how knowledge accumulation affects performance, we evaluate \OURS{} at different accumulation stages on \OURDATA{} with GPT-5.4-mini and Gemini-3-flash. As shown in Figure~\ref{fig:self_evolve_dynamics}, both Planning Accuracy and Unknown Tool Hit Rate improve consistently as the process advances.
These gains stem from the key mechanisms of the accumulation process, including contrastive experience extraction, reusable skill distillation, and dynamic tool migration.
Notably, both metrics grow rapidly from 0\% to 25\% progress and then improve more smoothly, suggesting that a limited amount of data is sufficient for the agent to establish robust priors for open-world adaptation. 
Detailed statistics are provided in Appendix~\ref{app:detailed_training_dynamics}.

\section{Conclusion}

In this work, we propose an ontology-aware self-evolving agent for open-world scientific tool acquisition.
\OURS{} follows a two-stage design: during accumulation, it distills reusable knowledge from contrastive trajectories; during inference, it uses active requests and a LinUCB-based bandit gate to balance tool exploration and exploitation. By leveraging an adaptive memory of skills, experiences, and an ontologized tool graph, \OURS{} enables agents to continually expand their capabilities in open-world scientific environments. Moreover, we introduce \OURDATA{}, a scientific tool-use benchmark with 900 tasks across four difficulty levels. Extensive experiments demonstrate that \OURS{} outperforms strong baselines, indicating its robustness and generalization.


\clearpage

\bibliography{aaai2027}

@article{ding2025scitoolagent,
  title={{SciToolAgent}: a knowledge-graph-driven scientific agent for multitool integration},
  author={Ding, Keyan and Yu, Jing and Huang, Junjie and Yang, Yuchen and Zhang, Qiang and Chen, Huajun},
  journal={Nature Computational Science},
  volume={5},
  number={10},
  pages={962--972},
  year={2025},
  publisher={Nature Publishing Group US New York}
}

@techreport{openai2025gpt5systemcard,
  title={{GPT-5} System Card},
  author={{OpenAI}},
  year={2025},
  month=aug,
  institution={OpenAI},
}

@article{yu2026scicueval,
  title={{SciCuEval}: {A} comprehensive dataset for evaluating scientific context understanding in large language models},
  author={Yu, Jing and Tang, Yuqi and Feng, Kehua and Liang, Lei and Zhang, Qiang and Ding, Keyan and Chen, Huajun},
  journal={Scientific Data},
  year={2026},
  publisher={Nature Publishing Group UK London}
}

@inproceedings{yao2023react,
  author = {Yao, Shunyu and Zhao, Jeffrey and Yu, Dian and Du, Nan and Shafran, Izhak and Narasimhan, Karthik R. and Cao, Yuan},
  title = {{ReAct}: Synergizing Reasoning and Acting in Language Models},
  booktitle = {The Eleventh International Conference on Learning Representations},
  year = {2023},
}

@article{shinn2023reflexion,
  title={Reflexion: Language agents with verbal reinforcement learning},
  author={Shinn, Noah and Cassano, Federico and Gopinath, Ashwin and Narasimhan, Karthik and Yao, Shunyu},
  journal={Advances in neural information processing systems},
  volume={36},
  pages={8634--8652},
  year={2023}
}

@article{wei2022chain,
  title={Chain-of-thought prompting elicits reasoning in large language models},
  author={Wei, Jason and Wang, Xuezhi and Schuurmans, Dale and Bosma, Maarten and Xia, Fei and Chi, Ed and Le, Quoc V and Zhou, Denny and others},
  journal={Advances in neural information processing systems},
  volume={35},
  pages={24824--24837},
  year={2022}
}

@article{yang2025qwen3,
  title={Qwen3 technical report},
  author={Yang, An and Li, Anfeng and Yang, Baosong and Zhang, Beichen and Hui, Binyuan and Zheng, Bo and Yu, Bowen and Gao, Chang and Huang, Chengen and Lv, Chenxu and others},
  journal={arXiv:2505.09388},
  year={2025}
}

@article{jiang2026xskill,
  title={Xskill: Continual learning from experience and skills in multimodal agents},
  author={Jiang, Guanyu and Su, Zhaochen and Qu, Xiaoye and Fung, Yi R},
  journal={arXiv:2603.12056},
  year={2026}
}

@inproceedings{li2010contextual,
  title={A contextual-bandit approach to personalized news article recommendation},
  author={Li, Lihong and Chu, Wei and Langford, John and Schapire, Robert E},
  booktitle={Proceedings of the 19th international conference on World wide web},
  pages={661--670},
  year={2010}
}

@article{schick2023toolformer,
  title={Toolformer: Language models can teach themselves to use tools},
  author={Schick, Timo and Dwivedi-Yu, Jane and Dess{\`\i}, Roberto and Raileanu, Roberta and Lomeli, Maria and Hambro, Eric and Zettlemoyer, Luke and Cancedda, Nicola and Scialom, Thomas},
  journal={Advances in neural information processing systems},
  volume={36},
  pages={68539--68551},
  year={2023}
}

@inproceedings{qin2024toolllm,
  title={{ToolLLM}: Facilitating large language models to master 16000+ real-world apis},
  author={Qin, Yujia and Liang, Shihao and Ye, Yining and Zhu, Kunlun and Yan, Lan and Lu, Yaxi and Lin, Yankai and Cong, Xin and Tang, Xiangru and Qian, Bill and others},
  booktitle={International Conference on Learning Representations},
  volume={2024},
  pages={9695--9717},
  year={2024}
}

@inproceedings{ma2024sciagent,
  title={Sciagent: Tool-augmented language models for scientific reasoning},
  author={Ma, Yubo and Gou, Zhibin and Hao, Junheng and Xu, Ruochen and Wang, Shuohang and Pan, Liangming and Yang, Yujiu and Cao, Yixin and Sun, Aixin},
  booktitle={Proceedings of the 2024 conference on empirical methods in natural language processing},
  pages={15701--15736},
  year={2024}
}

@article{min2023recent,
  title={Recent advances in natural language processing via large pre-trained language models: A survey},
  author={Min, Bonan and Ross, Hayley and Sulem, Elior and Veyseh, Amir Pouran Ben and Nguyen, Thien Huu and Sainz, Oscar and Agirre, Eneko and Heintz, Ilana and Roth, Dan},
  journal={ACM Computing Surveys},
  volume={56},
  number={2},
  pages={1--40},
  year={2023},
  publisher={ACM New York, NY}
}

@article{tang2025learning,
  title={Learning an Efficient Multi-Turn Dialogue Evaluator from Multiple LLM Judges},
  author={Tang, Yuqi and Feng, Kehua and Wang, Yunfeng and Chen, Zhiwen and Lv, Chengfei and Yu, Gang and Zhang, Qiang and Ding, Keyan and Chen, Huajun},
  journal={arXiv:2508.00454},
  year={2025}
}

@article{vemprala2024chatgpt,
  title={{ChatGPT} for robotics: Design principles and model abilities},
  author={Vemprala, Sai H and Bonatti, Rogerio and Bucker, Arthur and Kapoor, Ashish},
  journal={Ieee Access},
  volume={12},
  pages={55682--55696},
  year={2024},
  publisher={IEEE}
}

@inproceedings{li2023api,
  title={{Api-Bank}: A comprehensive benchmark for tool-augmented llms},
  author={Li, Minghao and Zhao, Yingxiu and Yu, Bowen and Song, Feifan and Li, Hangyu and Yu, Haiyang and Li, Zhoujun and Huang, Fei and Li, Yongbin},
  booktitle={Proceedings of the 2023 conference on empirical methods in natural language processing},
  pages={3102--3116},
  year={2023}
}

@article{song2023restgpt,
  title={{RestGPT}: Connecting large language models with real-world restful apis},
  author={Song, Yifan and Xiong, Weimin and Zhu, Dawei and Wu, Wenhao and Qian, Han and Song, Mingbo and Huang, Hailiang and Li, Cheng and Wang, Ke and Yao, Rong and others},
  journal={arXiv:2306.06624},
  year={2023}
}

@article{patil2024gorilla,
  title={Gorilla: Large language model connected with massive apis},
  author={Patil, Shishir G and Zhang, Tianjun and Wang, Xin and Gonzalez, Joseph E},
  journal={Advances in Neural Information Processing Systems},
  volume={37},
  pages={126544--126565},
  year={2024}
}

@inproceedings{wang2025toolgen,
  title={Toolgen: Unified tool retrieval and calling via generation},
  author={Wang, Renxi and Han, Xudong and Ji, Lei and Wang, Shu and Baldwin, Timothy and Li, Haonan},
  booktitle={International Conference on Learning Representations},
  volume={2025},
  pages={73473--73498},
  year={2025}
}

@inproceedings{trivedi2024appworld,
  title={Appworld: A controllable world of apps and people for benchmarking interactive coding agents},
  author={Trivedi, Harsh and Khot, Tushar and Hartmann, Mareike and Manku, Ruskin and Dong, Vinty and Li, Edward and Gupta, Shashank and Sabharwal, Ashish and Balasubramanian, Niranjan},
  booktitle={Proceedings of the 62nd Annual Meeting of the Association for Computational Linguistics (Volume 1: Long Papers)},
  pages={16022--16076},
  year={2024}
}

@inproceedings{chen2025scienceagentbench,
  title={Scienceagentbench: Toward rigorous assessment of language agents for data-driven scientific discovery},
  author={Chen, Ziru and Chen, Shijie and Ning, Yuting and Zhang, Qianheng and Wang, Boshi and Yu, Botao and Li, Yifei and Liao, Zeyi and Wei, Chen and Lu, Zitong and others},
  booktitle={International Conference on Learning Representations},
  volume={2025},
  pages={96934--96990},
  year={2025}
}

@inproceedings{mekala2024toolverifier,
  title={Toolverifier: Generalization to new tools via self-verification},
  author={Mekala, Dheeraj and Weston, Jason E and Lanchantin, Jack and Raileanu, Roberta and Lomeli, Maria and Shang, Jingbo and Dwivedi-Yu, Jane},
  booktitle={Findings of the Association for Computational Linguistics: EMNLP 2024},
  pages={5026--5041},
  year={2024}
}

@inproceedings{lu2025toolsandbox,
  title={Toolsandbox: A stateful, conversational, interactive evaluation benchmark for llm tool use capabilities},
  author={Lu, Jiarui and Holleis, Thomas and Zhang, Yizhe and Aumayer, Bernhard and Nan, Feng and Bai, Haoping and Ma, Shuang and Ma, Shen and Li, Mengyu and Yin, Guoli and others},
  booktitle={Findings of the Association for Computational Linguistics: NAACL 2025},
  pages={1160--1183},
  year={2025}
}

@article{team2023gemini,
  title={Gemini: a family of highly capable multimodal models},
  author={Team, Gemini and Anil, Rohan and Borgeaud, Sebastian and Alayrac, Jean-Baptiste and Yu, Jiahui and Soricut, Radu and Schalkwyk, Johan and Dai, Andrew M and Hauth, Anja and Millican, Katie and others},
  journal={arXiv:2312.11805},
  year={2023}
}

@article{qian2026toolrl,
  title={Toolrl: Reward is all tool learning needs},
  author={Qian, Cheng and Acikgoz, Emre Can and He, Qi and Wang, Hongru and Chen, Xiusi and Hakkani-Tur, Dilek and Tur, Gokhan and Ji, Heng},
  journal={Advances in Neural Information Processing Systems},
  volume={38},
  pages={105523--105553},
  year={2026}
}

@article{wang2025otc,
  title={{OTC}: Optimal tool calls via reinforcement learning},
  author={Wang, Hongru and Qian, Cheng and Zhong, Wanjun and Chen, Xiusi and Qiu, Jiahao and Huang, Shijue and Jin, Bowen and Wang, Mengdi and Wong, Kam-Fai and Ji, Heng},
  journal={arXiv--2504},
  year={2025}
}

@article{shen2023hugginggpt,
  title={Hugginggpt: Solving ai tasks with chatgpt and its friends in hugging face},
  author={Shen, Yongliang and Song, Kaitao and Tan, Xu and Li, Dongsheng and Lu, Weiming and Zhuang, Yueting},
  journal={Advances in Neural Information Processing Systems},
  volume={36},
  pages={38154--38180},
  year={2023}
}

@article{paranjape2023art,
  title={Art: Automatic multi-step reasoning and tool-use for large language models},
  author={Paranjape, Bhargavi and Lundberg, Scott and Singh, Sameer and Hajishirzi, Hannaneh and Zettlemoyer, Luke and Ribeiro, Marco Tulio},
  journal={arXiv:2303.09014},
  year={2023}
}

@inproceedings{liu2024agentbench,
  title={Agentbench: Evaluating llms as agents},
  author={Liu, Xiao and Yu, Hao and Zhang, Hanchen and Xu, Yifan and Lei, Xuanyu and Lai, Hanyu and Gu, Yu and Ding, Hangliang and Men, Kaiwen and Yang, Kejuan and others},
  booktitle={International Conference on Learning Representations},
  volume={2024},
  pages={52989--53046},
  year={2024}
}

@article{tang2023toolalpaca,
  title={Toolalpaca: Generalized tool learning for language models with 3000 simulated cases},
  author={Tang, Qiaoyu and Deng, Ziliang and Lin, Hongyu and Han, Xianpei and Liang, Qiao and Cao, Boxi and Sun, Le},
  journal={ arXiv:2306.05301},
  year={2023}
}

@article{m2024augmenting,
  title={Augmenting large language models with chemistry tools},
  author={M. Bran, Andres and Cox, Sam and Schilter, Oliver and Baldassari, Carlo and White, Andrew D and Schwaller, Philippe},
  journal={Nature machine intelligence},
  volume={6},
  number={5},
  pages={525--535},
  year={2024},
  publisher={Nature Publishing Group UK London}
}

@article{jin2024genegpt,
  title={{GeneGPT}: augmenting large language models with domain tools for improved access to biomedical information},
  author={Jin, Qiao and Yang, Yifan and Chen, Qingyu and Lu, Zhiyong},
  journal={Bioinformatics},
  volume={40},
  number={2},
  pages={btae075},
  year={2024},
  publisher={Oxford University Press}
}

@article{kang2024chatmof,
  title={{ChatMOF}: an artificial intelligence system for predicting and generating metal-organic frameworks using large language models},
  author={Kang, Yeonghun and Kim, Jihan},
  journal={Nature communications},
  volume={15},
  number={1},
  pages={4705},
  year={2024},
  publisher={Nature Publishing Group UK London}
}

@article{hess2009cognitive,
  title={Cognitive rigor: Blending the strengths of Bloom's taxonomy and Webb's depth of knowledge to enhance classroom-level processes},
  author={Hess, Karin and Jones, Ben and Carlock, Dennis and Walkup, John R},
  year={2009}
}

@article{lewis2020retrieval,
  title={Retrieval-augmented generation for knowledge-intensive nlp tasks},
  author={Lewis, Patrick and Perez, Ethan and Piktus, Aleksandra and Petroni, Fabio and Karpukhin, Vladimir and Goyal, Naman and K{\"u}ttler, Heinrich and Lewis, Mike and Yih, Wen-tau and Rockt{\"a}schel, Tim and others},
  journal={Advances in neural information processing systems},
  volume={33},
  pages={9459--9474},
  year={2020}
}

@inproceedings{ma2024m,
  title={m \& m’s: A benchmark to evaluate tool-use for m ulti-step m ulti-modal tasks},
  author={Ma, Zixian and Huang, Weikai and Zhang, Jieyu and Gupta, Tanmay and Krishna, Ranjay},
  booktitle={European Conference on Computer Vision},
  pages={18--34},
  year={2024},
  organization={Springer}
}

@article{wang2024survey,
  title={A survey on large language model based autonomous agents},
  author={Wang, Lei and Ma, Chen and Feng, Xueyang and Zhang, Zeyu and Yang, Hao and Zhang, Jingsen and Chen, Zhiyuan and Tang, Jiakai and Chen, Xu and Lin, Yankai and others},
  journal={Frontiers of Computer Science},
  volume={18},
  number={6},
  pages={186345},
  year={2024},
  publisher={Springer}
}

@article{wu2024chateda,
  title={Chateda: A large language model powered autonomous agent for eda},
  author={Wu, Haoyuan and He, Zhuolun and Zhang, Xinyun and Yao, Xufeng and Zheng, Su and Zheng, Haisheng and Yu, Bei},
  journal={IEEE Transactions on Computer-Aided Design of Integrated Circuits and Systems},
  volume={43},
  number={10},
  pages={3184--3197},
  year={2024},
  publisher={IEEE}
}

@article{chaudhari2026modular,
  title={Modular large language model agents for multi-task computational materials science},
  author={Chaudhari, Akshat and Ock, Janghoon and Barati Farimani, Amir and et al.},
  journal={Communications Materials},
  volume={7},
  number={1},
  pages={131},
  year={2026},
  publisher={Nature Publishing Group}
}

@article{ghareeb2026multi,
  title={A multi-agent system for automating scientific discovery},
  author={Ghareeb, Ali Essam and Chang, Benjamin and Mitchener, Ludovico and Yiu, Angela and Szostkiewicz, Caralyn J and Shved, Dmytro and Gyimesi, Gavin J and Laurent, Jon M and Wright, Samantha M and Razzak, Muhammed T and others},
  journal={Nature},
  pages={1--3},
  year={2026},
  publisher={Nature Publishing Group UK London}
}


\clearpage

\appendix

\setcounter{secnumdepth}{2}
\setcounter{table}{0}
\renewcommand{\thetable}{A\arabic{table}}
\renewcommand{\thesubsection}{\thesection\arabic{subsection}}
\raggedbottom

\section{Limitations}
Despite the efficacy of \OURS{} in open-world scientific tool acquisition, several limitations remain. First, our evaluation is conducted in a purely text-based setting, which abstracts away critical multi-modal aspects of real scientific workflows and leaves multi-modal tool adaptation underexplored.
For future work, we will evaluate our framework on multi-modal scientific tasks and explore dynamic memory updating and active unlearning strategies for more durable open-world tool adaptation.

\section{Additional Results}\label{app:additional_results}
In this section, we demonstrate the effectiveness of \OURS{} through additional experiments. We first analyze fine-grained ablations (Appendix~\ref{app:fine_grained_ablation}) and knowledge accumulation dynamics (Appendix~\ref{app:detailed_training_dynamics}). We then evaluate out-of-domain generalization (Appendix~\ref{app:ood_generalization}). Finally, we investigate inference efficiency (Appendix~\ref{app:inference_efficiency}). 

\subsection{Fine-Grained Ablation Study}\label{app:fine_grained_ablation}

In Section~\ref{sec:Ablation_Study}, we reported the overall component-level ablation results (Table~\ref{tab:ablation_study}). 
In this section, we conduct a fine-grained ablation of the ontology-aware inference pipeline using GPT-5.4-mini as the backbone to assess the contribution of each module. Table~\ref{tab:inference_ablation_study} yields the following observations: 
(1) Disabling the Task Router reduces Level-1 and Level-2 performance by 8.57 and 10.52 points, respectively, suggesting that it avoids unnecessary reasoning on simpler tasks. (2) Removing Active Request or replacing it with Retrieval Rewriting substantially degrades performance, indicating that requirement-conditioned retrieval better captures task-specific tool needs than raw-query matching. (3) Excluding Reranking degrades overall performance by 4.06 points, confirming the value of candidate refinement. (4) LinUCB outperforms both the Fixed Threshold Gate and the LLM-based Gate, indicating a more effective balance between known-tool exploitation and unknown-tool exploration. 
Overall, these results confirm that each component is necessary for an effective ontology-aware inference pipeline.

\begin{table*}[!t]
\centering
\footnotesize
\renewcommand{\arraystretch}{1}
\setlength{\tabcolsep}{3mm}
\resizebox{1.0\textwidth}{!}{
\begin{tabular}{lccccc}
\toprule
Method & Level-1 & Level-2 & Level-3 & Level-4 & Overall \\
\midrule
w/o Task Router & 77.14 \drop{8.57} & 66.32 \drop{10.52} & 61.18 \increase{0.37} & 36.84 \same{0.00} & 62.05 \drop{5.17} \\
w/o Active Request & 81.90 \drop{3.81} & 65.26 \drop{11.58} & 43.53 \drop{17.28} & 23.52 \drop{13.32} & 56.20 \drop{11.02} \\
Retrieval Rewriting & 82.86 \drop{2.85} & 66.32 \drop{10.52} & 49.41 \drop{11.40} & 26.32 \drop{10.52} & 58.73 \drop{8.49} \\
Query Decomposition & 82.86 \drop{2.85} & 76.84 \same{0.00} & 57.65 \drop{3.16} & 35.53 \drop{1.31} & 65.38 \drop{1.84} \\
w/o Reranking & 84.76 \drop{0.95} & 69.47 \drop{7.37} & 54.12 \drop{6.69} & 35.53 \drop{1.31} & 63.16 \drop{4.06} \\
Fixed Threshold Gate & 85.71 \same{0.00} & 74.74 \drop{2.10} & 58.82 \drop{1.99} & 30.26 \drop{6.58} & 64.82 \drop{2.40} \\
LLM-based Gate & 86.67 \increase{0.96} & 75.79 \drop{1.05} & 57.65 \drop{3.16} & 32.89 \drop{3.95} & 65.65 \drop{1.57} \\
\midrule
\rowcolor{modelgray}
\textbf{Ours - Full Pipeline} & 85.71 & 76.84 & 60.81 & 36.84 & 67.22 \\
\bottomrule
\end{tabular}
}
\caption{Fine-grained ablation study of ontology-aware inference modules on \textbf{\OURDATA}. All variants use GPT-5.4-mini as the backbone. We remove or replace key components to assess their contributions. \drop{} and \increase{} denote the absolute performance drop and improvement relative to the full pipeline, respectively, while gray values indicate no change.}
\label{tab:inference_ablation_study}
\end{table*}

\subsection{Detailed Knowledge Accumulation Statistics} \label{app:detailed_training_dynamics}
Table~\ref{tab:self_evolve_dynamics} presents detailed statistics of the knowledge accumulation process on \OURDATA{} with GPT-5.4-mini. 
We report results at five checkpoints, tracking the number of migrated unknown tools, the number of accumulated skills, the skill-based success rate, planning accuracy, and the hit rate on initially unknown tools.
As accumulation progresses, the numbers of acquired tools and accumulated skills increase steadily, while planning accuracy and the unknown-tool hit rate improve by 26.37 and 32.94 points, respectively. 
These results further demonstrate the importance of maintaining an evolving memory throughout knowledge accumulation.

\begin{table*}[!ht]
\centering
\renewcommand{\arraystretch}{1.2}
\begin{tabular}{lccccc}
\toprule
Progress & \#Acquired Tools & \#Skills & Skill Hit & Plan Acc. & Unk. Tool Hit \\
\midrule
0\%   & 0 & 0 & 1.20 & 47.25 & 55.30 \\
25\%  & 39 & 80 & 4.40 & 67.03 & 83.09 \\
50\%  & 62 & 147 & 7.69 & 68.13 & 85.29 \\
75\%  & 69 & 214 & 6.59 & 71.43 & 86.76 \\
100\% & 76 & 268 & 12.09 & 73.62 & 88.24 \\
\bottomrule
\end{tabular}
\caption{Performance evolution across knowledge accumulation stages on \textbf{\OURDATA} with GPT-5.4-mini. Metrics include the number of migrated tools (\#Acquired Tools), accumulated skills (\#Skills), Skill-based success rate (Skill Hit), planning accuracy (Plan Acc.), and hit rate on initially unknown tools (Unk. Tool Hit).}
\label{tab:self_evolve_dynamics}
\end{table*}

\subsection{Out-of-Domain Generalization}\label{app:ood_generalization}
While our primary focus remains on enhancing agent performance in open-world scientific tool acquisition, we also acknowledge the potential of \OURS{} in out-of-domain settings. To validate the generalizability of our approach, we conducted additional experiments on m\&m's~\cite{ma2024m}, a multi-step multimodal tool-use benchmark, and RestBench~\cite{song2023restgpt}, a benchmark for planning and executing RESTful API calls. Following the official evaluation protocols and metrics of each benchmark, we used GPT-5.4-nano as the backbone model for all methods and compared against the same baselines introduced in Section~\ref{sec:baselines} to ensure a fair comparison.

As shown in Table~\ref{tab:generalization_results}, \OURS{} achieves the highest tool-F1 (80.82) and argname-F1 (79.04) on m\&m's, outperforming prompting-based (Direct and CoT), retrieval-based (RAG and SciToolAgent), and interactive (ReAct and Reflexion) baselines. On RestBench, \OURS{} achieves the best overall performance on both TMDB and Spotify, particularly in success rate and completion percentage. Overall, these results demonstrate that \OURS{} generalizes effectively to broader tool-use domains.


\begin{table*}[!t]
\centering
\small
\renewcommand{\arraystretch}{1.05}
\setlength{\tabcolsep}{1.6mm}
\resizebox{\textwidth}{!}{
\begin{tabular}{lccccccc}
\toprule
\multirow{3}{*}{Method} & \multicolumn{2}{c}{m\&m's} & \multicolumn{5}{c}{RestBench} \\
\cmidrule(lr){2-3} \cmidrule(lr){4-8}
& \multirow{2}{*}{tool-F1 $\uparrow$} & \multirow{2}{*}{argname-F1 $\uparrow$} & \multicolumn{3}{c}{TMDB} & \multicolumn{2}{c}{Spotify} \\
\cmidrule(lr){4-6} \cmidrule(lr){7-8}
& & & Success\% $\uparrow$ & CP\% $\uparrow$ & $\Delta$ Sol. Len. $\downarrow$ & CP\% $\uparrow$ & $\Delta$ Sol. Len. $\downarrow$ \\
\midrule
Direct & 61.77 & 63.23 & 30.0 & 34.0 & 0.42 & 34.48 & 0.76 \\
CoT & 71.58 & 71.46 & \underline{44.0} & \underline{48.0} & $-0.06$ & 44.83 & $-0.07$ \\
RAG & \underline{71.62} & \underline{72.26} & 41.0 & 44.0 & $-0.10$ & 46.56 & $\mathbf{-0.36}$  \\
ReAct & 69.65 & 70.31 & 42.0 & 46.0 & 0.56 & 48.28 & 0.48 \\
Reflexion & 70.17 & 70.81 & 40.0 & 46.0 & 1.28 & \underline{51.72} & 0.83 \\
SciToolAgent & 65.48 & 67.92 & 40.0 & 42.0 & $\mathbf{-0.24}$ & 48.28 & \underline{$-0.14$} \\
\midrule
\textbf{SciToolAgent-Evo} & \textbf{80.82} & \textbf{79.04} & \textbf{54.0} & \textbf{62.0} & \underline{$-0.13$} & \textbf{53.45} & $\mathbf{-0.36}$ \\
\bottomrule
\end{tabular}
}
\caption{Comparison of model performance on \textbf{m\&m's} and \textbf{RestBench} using GPT-5.4-nano as the backbone. The best and second-best results in each column are bolded and underlined, respectively.}
\label{tab:generalization_results}
\end{table*}

\subsection{Inference Efficiency}\label{app:inference_efficiency}
We further compare the inference efficiency of \OURS{} with representative baselines on \OURDATA{} using GPT-5.4-mini. As shown in Table~\ref{tab:inference_cost}, although \OURS{} introduces additional overhead over SciToolAgent due to modules such as evolving-memory retrieval and active request formulation, it uses substantially fewer total tokens and has lower API cost than ReAct and Reflexion. It also achieves the lowest output-token usage and relatively low inference latency. Overall, \OURS{} maintains strong inference efficiency while enhancing open-world tool acquisition.

\begin{table*}[!t]
\centering
\small
\renewcommand{\arraystretch}{1.2}
\setlength{\tabcolsep}{1.5mm}
\resizebox{1.0\textwidth}{!}{
\begin{tabular}{lcccccc}
\toprule
Method & LLM Calls \textbf{$\downarrow$} & Input Tokens \textbf{$\downarrow$} & Output Tokens \textbf{$\downarrow$} & Total Tokens \textbf{$\downarrow$} & Latency (s) \textbf{$\downarrow$} & API Cost / Task (\$) \textbf{$\downarrow$} \\
\midrule
ReAct            & \underline{8.2}  & 72,115.85  & \underline{427.85} & 72,543.7  & 27.9 & 0.056 \\
Reflexion        & 16.1 & 149,435.95 & 931.15 & 150,367.1 & 44.7 & 0.116 \\
SciToolAgent     & \textbf{4.7} & \textbf{6,859.45} & 441.05 & \textbf{7,300.5} & \textbf{17.2} & \textbf{0.007} \\
\midrule
\textbf{SciToolAgent-Evo} & 11.3 & \underline{21,711.35} & \textbf{387.8} & \underline{22,099.15} & \underline{26.4} & \underline{0.018} \\
\bottomrule
\end{tabular}
}
\caption{Inference cost comparison on \textbf{\OURDATA} with GPT-5.4-mini. All reported results are averaged across tasks, with the best result bolded and the second-best underlined.}
\label{tab:inference_cost}
\end{table*}




\section{Task Levels and Prompts}\label{app:generation_prompts}
In this section, we provide detailed definitions of the four levels in \OURDATA\ and present the prompt templates below. The distinctions among these levels are inspired by the Cognitive Rigor Matrix~\cite{hess2009cognitive}, with task difficulty characterized jointly by knowledge mapping, reasoning depth, constraint integration, and decision complexity.

\textbf{Level-1 (1--2 tools)} consists of explicit invocation tasks, where the question explicitly provides the tool name or clear hints to the required tool. These tasks typically involve a single tool call or a short tool chain and focus on basic data retrieval for a single scientific entity.

\vspace{0.3em}
\begin{prompt}{Prompt for Generating Level-1 Questions}
\textbf{System Prompt:}

You are a science expert.

\bigskip
\textbf{User Prompt:}

Please generate one scientific tool-use question based on the given candidate tools. The question type should be one of Question \& Answer, Multiple-Choice, Content Completion, and True/False.

\bigskip
\textbf{Candidate Tools:}

\{tools\_str\}

\bigskip
\textbf{Task Objective:}

Generate a basic instruction-level task. The question should focus on a single substance and use standard scientific terminology to clearly indicate the capability corresponding to a single tool or a very short tool chain.

\bigskip
\textbf{[RULE]:}

1. The question must evaluate explicit instruction following and clearly require a specific scientific operation.

2. tool\_path should contain a single tool or a very short tool chain arranged in the correct execution order.

3. The question may not describe the solving steps or reveal any tool outputs, numerical results, or answer hints.

4. The Parameter field should contain only the substance name or ID selected in the question.

5. Ensure natural, diverse phrasing and avoid template-like wording.

Strictly output the response in \textbf{valid JSON} format only.
The required \textbf{JSON} format is as follows:

\{ \\
 \hspace*{0.5em} "tool\_path": ["tool\_selected", "xxx"], \\
 \hspace*{0.5em} "question": "question text",  \\
 \hspace*{0.5em} "Parameter": "substance name" \\
\}
\end{prompt}
\vspace{0.3em}

\textbf{Level-2 (2--4 tools)} consists of property analysis tasks, where the question focuses on a single or a small number of target properties. These tasks typically involve short tool chains for property retrieval, direct computation, or comparison, requiring the model to select appropriate tools and make direct judgments based on their outputs.

\vspace{0.3em}
\begin{prompt}{Prompt for Generating Level-2 Questions}
\textbf{System Prompt:}

You are a science expert.

\bigskip
\textbf{User Prompt:}

Please generate one scientific tool-use question based on the given candidate tools. The question type should be one of Question \& Answer, Multiple-Choice, Content Completion, and True/False.

\bigskip
\textbf{Candidate Tools:}

\{tools\_str\}

\bigskip
\textbf{Task Objective:}

Generate a terminology-level task. The question should use standard scientific terminology to imply the target property to be evaluated. It typically involves a short tool chain from entity representation conversion to property computation, comparison, or filtering over a few substances, requiring the model to infer the necessary tool capabilities from the property requirements.

\bigskip
\textbf{[RULE]:}

1. The question must evaluate implicit grounding from scientific terminology to a tool-capability chain.

2. tool\_path should contain a short tool chain arranged in the correct execution order, supporting entity representation conversion, property computation, or property comparison.

3. The question must not directly mention any tool name, but should use standard scientific terminology to clearly imply the target property to be evaluated; it must not describe the solving steps or reveal any tool outputs, numerical results, or answer hints.

4. The Parameter field should contain only the referenced substance names or IDs, with a consistent format.

5. The question must not use comparison criteria that are unnatural in real scientific queries, such as median; instead, it should use more realistic scientific constraints such as maximum/minimum values, threshold-based judgments, or structural inclusion.

6. Ensure natural, diverse phrasing and avoid template-like wording.

\bigskip
Strictly output the response in \textbf{valid JSON} format only.
The required \textbf{JSON} format is as follows:

\{ \\
 \hspace*{0.5em} "tool\_path": ["tool\_selected", "xxx"], \\
 \hspace*{0.5em} "question": "question text",  \\
 \hspace*{0.5em} "Parameter": ["substance name 1", "substance name 2", "xxx"] \\
\}
\end{prompt}
\vspace{0.3em}

\textbf{Level-3 (3--6 tools)} consists of constraint reasoning tasks, where the question specifies multiple candidate entities, multi-dimensional interrelated properties, and explicit scientific constraints. These tasks require integrating information from multiple sources and performing constraint-aware reasoning to infer latent relationships and identify the most appropriate candidate.

\vspace{0.3em}
\begin{prompt}{Prompt for Generating Level-3 Questions}
\textbf{System Prompt:}

You are a science expert.

\bigskip
\textbf{User Prompt:}

Please generate one scientific tool-use question based on the given candidate tools. The question type should be one of Question \& Answer, Multiple-Choice, Content Completion, and True/False.

\bigskip
\textbf{Candidate Tools:}

\{tools\_str\}

\bigskip
\textbf{Task Objective:}

Generate an expert decision-level task. The question should provide multiple candidate objects and multiple scientific constraints. It typically requires cross-tool and cross-attribute integration over multiple substances, requiring the model to perform candidate selection and execute a relatively long tool chain under multiple constraints.

\bigskip
\textbf{[RULE]:}

1. The question must evaluate deep scientific reasoning, including property derivation, comprehensive threshold-based judgment, candidate filtering, or selection of the optimal object satisfying all conditions.

2. tool\_path should contain a multi-tool chain arranged in the correct execution order, supporting cross-attribute analysis and constraint-based reasoning over multiple candidate objects.

3. The question must not directly mention any tool name, but should use standard scientific terminology to clearly describe the candidate objects, target properties, and scientific constraints; it must not describe the solving steps or reveal any tool outputs, numerical results, or answer hints.

4. The Parameter field should contain only the candidate substance names or IDs involved in the question, with a consistent representation format.

5. The question must not use comparison criteria that are unnatural in real scientific queries, such as median; instead, it should use scientific constraints such as maximum/minimum values, threshold-based judgments, or structural inclusion.

6. Ensure natural, diverse phrasing and avoid template-like wording.

\bigskip
Strictly output the response in \textbf{valid JSON} format only.
The required \textbf{JSON} format is as follows:

\{ \\
 \hspace*{0.5em} "tool\_path": ["tool\_selected", "xxx"], \\
 \hspace*{0.5em} "question": "question text",  \\
 \hspace*{0.5em} "Parameter": ["candidate substance 1", "candidate substance 2", "xxx"] \\
\}
\end{prompt}
\vspace{0.3em}

\textbf{Level-4 (5--10 tools)} consists of research strategy tasks, where the question simulates realistic scientific scenarios with open-ended research goals and multi-dimensional evaluation criteria. These tasks typically involve long tool chains and multi-objective trade-offs, requiring the model to autonomously plan a complete analysis workflow and formulate interpretable research decisions.

\vspace{0.3em}
\begin{prompt}{Prompt for Generating Level-4 Questions}
\textbf{System Prompt:}

You are a science expert.

\bigskip
\textbf{User Prompt:}

Please generate one scientific tool-use question based on the given candidate tools. The question type should be one of Question \& Answer, Multiple-Choice, Content Completion, and True/False.

\bigskip
\textbf{Candidate Tools:}

\{tools\_str\}

\bigskip
\textbf{Task Objective:}

Generate a research strategy-level task. The question should simulate a realistic scientific scenario, specify an open-ended research objective and multidimensional evaluation criteria, and typically involve a long tool chain and multi-objective trade-offs, requiring the model to autonomously plan the analysis workflow and produce a rigorous scientific decision.

\bigskip
\textbf{[RULE]:}

1. The question must present a complete and realistic scientific motivation, avoiding a rote list of tool calls; it may focus on drug screening, materials selection, property prediction, mechanism investigation, safety assessment, or de novo design validation.

2. The question should use 1-2 sentences to establish the research background, followed by precise and non-redundant scientific statements specifying the analysis objective and evaluation criteria.

3. tool\_path should contain a relatively long tool chain arranged in a reasonable execution order, supporting multidimensional analysis, evidence integration, and scientific decision-making.

4. The question must not directly mention any tool name, but should use standard scientific terminology to describe the research background, analysis objective, and evaluation criteria; it must not describe the specific solving steps or reveal any tool outputs, numerical results, or answer hints.

5. The Parameter field should contain the specific substances, sequences, or other scientific objects selected in the question.

6. Ensure natural, coherent, and situated in a realistic research context.

\bigskip
Strictly output the response in \textbf{valid JSON} format only.
The required \textbf{JSON} format is as follows:

\{ \\
 \hspace*{0.5em} "tool\_path": ["tool\_selected", "xxx"], \\
 \hspace*{0.5em} "question": "question text",  \\
 \hspace*{0.5em} "Parameter": ["scientific object 1", "scientific object 2", "xxx"] \\
\}
\end{prompt}
\vspace{0.3em}

\begin{table*}[!ht]
\centering
\small
\begin{tabular}{lcccccccccccccc}
\toprule
Level & \#Questions & \#Train & \#Test & Q\&A & MCQ & CC & T/F 
& \makecell{Avg.\\Path} 
& \makecell{Max.\\Path} 
& \makecell{Avg.\\\#Param.} 
& \makecell{Avg.\\\#Cons.} 
& \makecell{Avg.\\\#Attr.} 
& \makecell{Avg.\\Depth} 
& \#Reuse \\
\midrule
Level-1 & 262 & 157 & 105 & 253 & 0 & 4 & 5  & 1.57 & 2  & 1.00 & 0.00 & 1.14 & 1.00 & 54 \\
Level-2 & 238 & 143 & 95 & 63 & 79 & 31 & 65 & 2.26 & 4  & 3.23 & 1.84 & 1.45 & 2.18 & 31 \\
Level-3 & 213 & 128 & 85 & 39 & 70 & 63 & 41 & 3.68 & 6  & 4.11 & 3.25 & 2.61 & 3.52 & 26 \\
Level-4 & 187 & 111 & 76 & 89 & 35 & 28 & 35 & 5.21 & 10 & 1.61 & 4.76 & 4.12 & 4.88 & 17 \\
\midrule
Total   & 900 & 539 & 361 & 444 & 184 & 126 & 146 & 3.03 & 10 & 2.45 & 2.24 & 2.19 & 2.71 & 128 \\
\bottomrule
\end{tabular}
\caption{Statistics of the \textbf{\OURDATA} dataset across four difficulty levels, covering dataset scale, question formats, tool-path structure, and task complexity. 
\textit{Avg. Path} and \textit{Max. Path} denote the average and maximum tool-path length;
\textit{Avg. \#Param.}, \textit{Avg. \#Cons.}, and \textit{Avg. \#Attr.} denote the average number of input parameters, constraints, and tool attributes per question; 
\textit{Avg. Depth} denotes the average reasoning depth, and \textit{\#Reuse} denotes the number of reusable skills.}
\label{tab:dataset_statistics}
\end{table*}

\section{Benchmark Validation}\label{app:benchmark_validation}
This section details the benchmark validation protocol of \OURDATA, which consists of two stages, namely LLM-as-a-Judge verification and human validation~\cite{yu2026scicueval}.

\textbf{LLM-as-a-Judge.} 
We used Gemini-3.1-Pro as the judge to assess each generated sample along three dimensions: (1) whether the question is clearly stated and free of obvious ambiguity; (2) whether the reference tool chain fully covers the scientific operations required by the question; and (3) whether the input-output dependencies between adjacent tools in the chain are valid.  
Samples that failed this verification process were removed.

\vspace{0.3em}
\begin{prompt}{Prompt for Sample Quality Evaluation}
\textbf{System Prompt:}

You are a scientific evaluation expert.

\bigskip
\textbf{User Prompt:}

Below are a question, a reference tool chain, and the corresponding parameter. Please use the following rule to determine whether the sample satisfies the quality standard.

\bigskip
\textbf{[RULE]:}

1. Check whether the question is clearly stated and free of obvious ambiguity.

2. Check whether tool\_path covers the scientific operations required by the question.

3. Check whether the input-output dependencies between adjacent tools in tool\_path are valid and can form an executable chain.


4. Judge as ``No'' if any condition fails; otherwise, ``Yes''.

\bigskip
\textbf{Question:}

\{question\}

\bigskip
\textbf{Reference tool chain:}

\{tool\_path\}

\bigskip
\textbf{Parameter:}

\{parameter\}

\bigskip
Strictly output the response in \textbf{valid JSON} format only. 
The required \textbf{JSON} format is as follows:

\{ \\
 \hspace*{0.5em} "decision": "Yes / No", \\
 \hspace*{0.5em} "reason": "brief explanation" \\
\}
\end{prompt}
\vspace{0.3em}

\textbf{Human Validation.}
To further ensure the quality and rigor of our dataset, we conducted human validation on the samples that passed the initial LLM-based verification. For Levels 1--3, we invited five students with backgrounds in biology, chemistry, and materials science to perform the review. For Level-4, we invited three domain experts, whose expertise was better suited for evaluating the open-ended research scenarios. Each sample was independently reviewed by three annotators according to the following four criteria:

\begin{itemize}[left=0pt]
    \item \textbf{Level Alignment}: whether the question effectively tests the design objective of the assigned level and accurately reflects the core competency intended to be evaluated at that level.
    \item \textbf{Semantic Clarity}: whether the question is clearly stated, logically coherent, and free of obvious ambiguity, such that the scientific goal and constraints can be accurately understood.
    \item \textbf{Tool-Chain Coherence}: whether the reference tool chain can fully support the question, and whether the input-output dependencies between consecutive tools are valid, avoiding mismatched tool capabilities or broken data flows.
    \item \textbf{Answerability}: whether the involved scientific entities exist, and whether the answer can be obtained through executing the given tool chain and applying the necessary model reasoning.
\end{itemize}

Student reviewers were compensated at a rate of \$10 for every 20 cases, while expert reviewers received \$15 for every 10 cases. The entire validation process took five days and cost \$1,900 in total. Each reviewer independently provided a ``Yes'' or ``No'' judgment for every assigned sample. We then applied a majority-vote rule to aggregate the three reviewers' assessments and obtain the final consensus decision. Among the samples entering human validation, 85.7\% satisfied the above criteria, resulting in a final benchmark of 900 questions across four difficulty levels, three scientific domains, and four question formats. 
Detailed dataset statistics are reported in Tables~\ref{tab:dataset_statistics}.

\section{Ontology-Aware Inference Algorithm}\label{app:algorithm1}
Algorithm~\ref{alg:active_tool_acquisition} presents the details of ontology-aware inference, including skill and task routing, active tool acquisition, and knowledge update.

\begin{algorithm}[!t]
\caption{Ontology-Aware Inference}\label{alg:active_tool_acquisition}
\small
\begin{algorithmic}[1]
\REQUIRE Test task $q$, task domain $d_q$, memory $\mathcal{M}=(G_K,\mathcal{K},\mathcal{E})$, unknown tools $T_U$, bandit gate $\mathcal{B}$
\ENSURE Answer $\hat{y}$, updated $\mathcal{M}$, $T_U$ and $\mathcal{B}$
\STATE $\hat{\tau}\gets[\,]$, $\mathcal{A}\gets[\,]$

\vspace{0.3em}
\STATE \textcolor{blue!80}{\textbf{Step 1: Skill and task routing}}
\STATE $k^\star \gets \textsc{SkillMatch}(q,\mathcal{K})$
\IF{$k^\star \neq \emptyset$}
    \STATE $\hat{\tau} \gets$ Tool chain stored in $k^\star$
    \STATE $\hat{y} \gets \textsc{Execute}(q,\hat{\tau})$
    \STATE \textbf{return} $\hat{y},\mathcal{M},T_U,\mathcal{B}$
\ENDIF
\STATE $c_q \gets \textsc{AssessDifficulty}(q)$
\IF{$c_q=\textsc{Simple}$}
    \STATE $\hat{\tau} \gets$ CoT planning over $G_K \cup T_U$
    \STATE $\hat{y} \gets \textsc{Execute}(q,\hat{\tau})$
    \STATE \textbf{return} $\hat{y},\mathcal{M},T_U,\mathcal{B}$
\ENDIF

\vspace{0.3em}
\STATE \textcolor{blue!80}{\textbf{Step 2: Active Tool acquisition}}
\STATE $\mathcal{E}_q \gets \mathcal{E}_{gen} \cup \textsc{TopK}(\mathcal{E}_{d_q},q)$
\STATE $R_q=\{r_1,\ldots,r_l\} \gets \textsc{ActiveRequest}(q,\mathcal{E}_q)$
\FOR{each request $r_j\in R_q$}
    \STATE $C_j^0 \gets \textsc{TopK}_{t\in V_K} S(r_j,t)$
    \STATE $C_j \gets \textsc{SubgraphExpand}(C_j^0,G_K)$
    \STATE $x_j=[f_j,m_j,s_j,p_j]^\top \gets \textsc{GateCtx}(r_j,C_j)$
    \STATE $a_j \gets \mathcal{B}(x_j)$, where $a_j\in\{\textsc{Known},\textsc{Unknown}\}$
    \IF{$a_j=\textsc{Known}$}
        \STATE $t_j \gets \textsc{Top}(C_j)$
    \ELSE
        \STATE $t_j \gets \textsc{SearchUnknown}(r_j,T_U)$
    \ENDIF
    \STATE $\hat{\tau}\gets \hat{\tau}\oplus t_j$, \quad $\mathcal{A}\gets \mathcal{A}\oplus (x_j,a_j)$
\ENDFOR

\vspace{0.3em}
\STATE \textcolor{blue!80}{\textbf{Step 3: Knowledge update}}
\STATE $(\hat{y},H,r)\gets \textsc{Execute}(q,\hat{\tau})$, where $r\in\{0,1\}$
\IF{$r=1$}
    \FOR{each unknown tool $u\in \hat{\tau}$}
        \STATE $o_u \gets \textsc{CompleteOntology}(u,H)$
        \STATE $G_K \gets \textsc{Migrate}(G_K,u,o_u)$
        \STATE $T_U \gets T_U\setminus \{u\}$
    \ENDFOR
    \STATE $\mathcal{K}\gets \mathcal{K}\cup \textsc{SkillDistill}(q,\hat{\tau},H)$
\ENDIF
\STATE $\mathcal{M}\gets(G_K,\mathcal{K},\mathcal{E})$
\STATE \textbf{return} $\hat{y},\mathcal{M},T_U,\mathcal{B}$
\end{algorithmic}
\end{algorithm}
\normalsize

\section{Active Tool Request Analysis}\label{app:active_request}

In this section, we provide a brief analysis of why \OURS{} adopts active tool requests instead of passive tool selection.

\noindent Let $\bar{\mathcal{T}}=\mathcal{T}_K\cup\tilde{\mathcal{T}}_U$ denote the conditionally observable tool space, where $\mathcal{T}_K$ is the known tool graph and $\tilde{\mathcal{T}}_U$ contains unknown-tool attributes accessed only when $G_K$ is insufficient. Let $\Pi(\bar{\mathcal{T}})$ denote the candidate tool-chain space and $\tau^*$ the reference tool chain for task $q$.

\noindent A passive strategy directly matches the raw task against candidate tool chains:
\begin{equation}\label{app_eq:1}
\resizebox{0.89\linewidth}{!}{$
\begin{aligned}
P_{\mathrm{passive}}(\tau^*\mid q,\bar{\mathcal{T}})
=
\frac{
P(q\mid \tau^*,\bar{\mathcal{T}})P(\tau^*\mid \bar{\mathcal{T}})
}{
\sum_{\tau\in\Pi(\bar{\mathcal{T}})}
P(q\mid \tau,\bar{\mathcal{T}})P(\tau\mid \bar{\mathcal{T}})
}.
\end{aligned}
$}
\end{equation}
This direct query-based matching is inefficient in large heterogeneous tool spaces and unreliable for weakly described unknown tools.

\noindent In contrast, given $q$ and retrieved experiences $\mathcal{E}_q$, \OURS{} actively generates a request set $R_q=\{r_1,r_2,\ldots,r_l\}$, where each $r_j$ is a concise description of one tool need. Tool-chain acquisition is then conditioned on these requests:
\begin{equation}\label{app_eq:2}
\resizebox{0.89\linewidth}{!}{$
\begin{aligned}
P_{\mathrm{active}}(\tau^*\mid q,\mathcal{E}_q,\bar{\mathcal{T}})
=
\sum_{R_q}
P(\tau^*\mid R_q,\bar{\mathcal{T}})
P(R_q\mid q,\mathcal{E}_q).
\end{aligned}
$}
\end{equation}
Here, $P(R_q\mid q,\mathcal{E}_q)$ captures the agent's ability to express task-specific tool needs under ontology-guided reasoning.

From an information-acquisition view, an effective request set should reduce uncertainty over the required tool chain:
\begin{equation}\label{app_eq:3}
\resizebox{0.89\linewidth}{!}{$
\begin{aligned}
R_q^*
&=
\arg\max_{R_q}
I(\tau^*;R_q\mid q,\mathcal{E}_q) \\
&=
\arg\max_{R_q}
\left[
H(\tau^*\mid q,\mathcal{E}_q)
-
H(\tau^*\mid R_q,q,\mathcal{E}_q)
\right].
\end{aligned}
$}
\end{equation}
Moreover, functionality-level requests are better aligned with both known-tool ontology and unknown-tool functionalities than raw task queries, since they describe the required capability in the same semantic form as tool descriptions. Thus, active tool requests transform scientific tool acquisition from costly raw-query matching into requirement-conditioned retrieval.

\section{Experimental Settings}\label{app:experimental_configuration}

During knowledge accumulation, we first sample a deterministic baseline trajectory ($T_{\mathrm{base}} = 0$). If this trajectory fails, we further sample three exploratory trajectories in parallel with $T_{\mathrm{exp}}\in\{0.2,0.5,0.8\}$. After accumulation, we deduplicate the experience pool by removing experience rules whose pairwise semantic similarity exceeds $\theta_{\mathrm{sim}}$ = 0.85. 

At inference time, the agent utilizes all global experience rules and retrieves the top-5 domain-specific entries. The decoding temperature of the Main Agent is set to 0.3, with max retries = 2, and the final answer is evaluated by GPT-5-mini. We split the toolset into two equal partitions, designating 50\% as known and the remaining 50\% as unknown. The difficulty routing module uses two branch thresholds, namely an estimated tool count of 2 and a scientific entity count of 4. During this stage, locally deployed Qwen3-Embedding-0.6B is used as the retrieval model ($max\_tools = 5$), while Qwen3-1.7B is used as the reranking model.  We implement a LinUCB-based bandit gate with a 4D context vector. The exploration coefficient $\alpha$ initializes at $0.8$ and decays by $0.01$ per step to a minimum of $0.2$. Furthermore, a 20-task sliding window monitors the known-tool failure rate, dynamically rescaling $\alpha$ within $[0.7, 1.3]$ to adaptively balance exploration and exploitation.

Specifically, we adopt a unified and standardized downstream pipeline for tool execution, summarization, and evaluation. This setup ensures a fair comparison by confining methodological variations strictly to the tool-chain planning strategies.

\section{Introduction of Baselines}\label{app:baseline_details}
We have selected the following methods as our baselines:

\begin{itemize}[left=0pt]
    \item \textbf{Direct}: Generates the final result directly without intermediate reasoning trajectories. In our setting, the model first identifies suitable tools from the known set, retrieves candidates from the unknown set if necessary, and directly outputs the planned tool chain for execution and summarization.
    \item \textbf{CoT}~\cite{wei2022chain}: Produces an explicit reasoning process prior to generating the result. Specifically, it reasons over the known and unknown tool sets sequentially to identify required tools, subsequently merging them into a coherent chain for execution.
    \item \textbf{RAG}~\cite{lewis2020retrieval}: Augments tool-chain planning with retrieved tool candidates. It retrieves relevant tools from the known graph and, when capability gaps are identified, retrieves additional candidates from the unknown set before constructing the tool chain for execution.
    \item \textbf{ReAct}~\cite{yao2023react}: Integrates reasoning with action execution through an iterative Thought-Action-Observation loop. The model generates a reasoning trajectory to select and execute actions, dynamically updating its plan and exploring unknown tools based on real-time observations until completion.
    \item \textbf{Reflexion}~\cite{shinn2023reflexion}: Improves agent planning through verbal self-reflection. When an attempt fails, the model analyzes the failure mode to generate feedback, which then informs and refines subsequent planning attempts.
    \item \textbf{SciToolAgent}~\cite{ding2025scitoolagent}: Leverages a scientific tool knowledge graph to facilitate multi-tool orchestration. It retrieves relevant tools from the graph and searches the unknown set when gaps are identified, following which it synthesizes and executes the tool chain.
\end{itemize}



\section{Detailed Prompts for Agent Workflow}\label{app:workflow_prompts}

To ensure the robustness and reproducibility of \OURS, we present the core prompt templates used in our workflow, covering active tool requirement description, candidate tool reranking and scoring, experience rule generation, and reusable skill extraction. 

\vspace{0.3em}
\begin{prompt}{Prompt for Active Tool Description}
\textbf{System Prompt:}

You are a science expert.

\bigskip
\textbf{User Prompt:}

You need to describe the tool capabilities required for a scientific task.

\bigskip
\textbf{Experience:}
\{experience\_section\}

\bigskip
\textbf{Task:}

\{question\}

\bigskip
\textbf{Previous Error (if any):}

\{error\_text\}

\bigskip
\textbf{[IMPORTANT RULE]:}

1. Each tool capability must be an atomic computational primitive, used only to derive one property or perform one transformation. It must not merge multiple properties or operations, nor include decision logic such as comparison, filtering, or aggregation.

2. Use the fewest and most direct steps for task analysis. If a single tool capability can directly produce the target result, do not introduce redundant preprocessing, intermediate transformations, or repeated computations. Each type of property should be computed only once.

3. If the task involves state or representation transformation, tool capabilities must follow data-form evolution, ensuring causal coherence between steps without skipping intermediate forms.

4. Tool descriptions should be generic capability descriptions and must not include specific substance names, numerical values, candidate answers, task conclusions, or instantiated information.

5. If the task involves multiple substances, sequences, or scientific objects, describe only the tool capabilities required for processing one object; do not introduce cross-object comparison or selection logic.

6. Use goal-oriented backward reasoning by first determining the final output internally, then inferring the required inputs and tool capabilities, and finally outputting only the tool capability descriptions.

7. Each tool description must be highly concise, with no more than 15 English words.

\bigskip
Now analyze the given task and list all required tool capabilities.

\bigskip
\textbf{The required output format is as follows:}

\textit{\hspace*{0.5em} (1) tool description 1 \\
\hspace*{0.5em} (2) tool description 2 \\
\hspace*{1.0em} ...}

\bigskip
Do not output any additional explanation.
\end{prompt}
\vspace{0.3em}


\vspace{0.3em}
\begin{prompt}{Prompt for Rerank and Score}
\textbf{System Prompt:}

You are a science expert.

\bigskip
\textbf{User Prompt:}

Complete the following two-step task.

\bigskip
\textbf{Task Requirement:}

\{tool\_request\}

\bigskip
\textbf{Task Constraints:}

Category=\{target\_category\}\{expected\_input\}

\bigskip
\textbf{Candidate Tools:}

\{candidates\_text\}

\bigskip
\textbf{Step 1:}

Analyze each candidate tool based on its Function description, Input type, Output type, and Category. Select the best-matching tool as top-1 and the second-best tool as top-2.

\bigskip
\textbf{Step 2:}

Score the selected top-1 and top-2 tools according to the following criteria:

\bigskip
\textbf{1. known\_top1\_feasibility} $(0.0$--$1.0)$: the overall feasibility of top-1.

\begin{itemize}[leftmargin=*]
    \item \textbf{\textit{$1.0$}}: Function fully matches, with aligned I/O and Category.
    \item \textbf{\textit{$0.75$}}: Function strongly matches, with minor I/O or Category mismatch.
    \item \textbf{\textit{$0.5$}}: Function partially matches, with moderate I/O or Category mismatch.
    \item \textbf{\textit{$0.25$}}: Function is weakly relevant, with clear I/O or Category mismatch.
    \item \textbf{\textit{$0.0$}}: Function is completely mismatched.
\end{itemize}

\textbf{2. known\_margin}: the rationale for prioritizing top-1 over top-2.

\begin{itemize}[leftmargin=*]
    \item \textbf{\textit{$1.0$}}: top-1 is clearly better than top-2.
    \item \textbf{\textit{$0.75$}}: top-1 is slightly better than top-2.
    \item \textbf{\textit{$0.5$}}: top-1 and top-2 are comparable.
    \item \textbf{\textit{$0.25$}}: top-2 is slightly better than top-1.
    \item \textbf{\textit{$0.0$}}: top-2 is clearly better than top-1.
\end{itemize}

\textbf{3. ontology\_alignment\_score} $(0.0$--$1.0)$: the overall ontology-level alignment between top-1 and the task.

\begin{itemize}[leftmargin=*]
    \item \textbf{\textit{$1.0$}}: Category and I/O types are fully aligned.
    \item \textbf{\textit{$0.75$}}: Category matches, with minor I/O type mismatch.
    \item \textbf{\textit{$0.5$}}: Category matches but I/O types differ, or I/O matches but Category differs.
    \item \textbf{\textit{$0.25$}}: Category and I/O both differ but remain partially relevant.
    \item \textbf{\textit{$0.0$}}: Category or I/O types are completely mismatched.
\end{itemize}

Strictly output the response in \textbf{valid JSON} format only.
The required \textbf{JSON} format is as follows:

\{ \\
 \hspace*{0.5em} "top1": "ToolName1", \\
 \hspace*{0.5em} "top2": "ToolName2", \\
 \hspace*{0.5em} "known\_top1\_feasibility": F, \\
 \hspace*{0.5em} "known\_margin": M, \\
 \hspace*{0.5em} "ontology\_alignment\_score": O \\
\}
\end{prompt}
\vspace{0.3em}

\vspace{0.3em}
\begin{prompt}{Prompt for Generating Experience Rules}
\textbf{System Prompt:}

You are a science expert.

\bigskip
\textbf{User Prompt:}

Based on the diagnosis and the rules, generate highly reusable experience rules.

\bigskip
\textbf{Question:}

\{question\}

\bigskip
\textbf{Diagnosis:}

\{diagnosis\}

\bigskip
Please generate the following fields:

\bigskip
\textbf{1. general\_prompt:} One general principle applicable to scientific tool planning. \\
\textbf{2. domain:} Strictly choose one from the following: biology, chemistry, materials. \\
\textbf{3. domain\_prompt:} A robust tool mapping rule targeting the diagnosed error.

\bigskip
The core expression must follow:

\textit{``When requiring [specific scientific context/constraint], select a tool that provides [specific functional capability].''}

\bigskip
\textbf{[RULE]:}

1. All rules must provide clear and actionable operational guidance.

2. Limit each rule to 25 words.

\bigskip
Strictly output the response in \textbf{valid JSON} format only.
The required \textbf{JSON} format is as follows:

\{ \\
 \hspace*{0.5em} "general\_prompt": "1 universal actionable rule", \\
 \hspace*{0.5em} "domain": "biology/chemistry/materials", \\
 \hspace*{0.5em} "domain\_prompt": "domain-specific rule" \\
\}
\end{prompt}
\vspace{0.3em}

\vspace{0.3em}
\begin{prompt}{Prompt for Generating Reusable Skills}
\textbf{System Prompt:}

You are a science expert.

\bigskip
\textbf{User Prompt:}

Given a specific task and its successfully executed tool chain, extract one standardized and reusable skill.

\bigskip
\textbf{Task:}

\{question\}

\bigskip
\textbf{Tool Chain:}

\{tool\_path\}

\bigskip
\textbf{Core Extraction Criteria:}

1. Avoid over-abstraction: preserve specific scientific properties.

2. Remove concrete parameters: do not include numerical values, thresholds, filtering conditions, or specific substance names.

3. Skill naming: the key Ni should use short snake\_case and reflect only the property/function combination.

4. Skill description: the key Zi must be a coherent sequence of atomic computational primitives. Use connectives to form one fluent description within 20 words.

\bigskip
Strictly output the response in \textbf{valid JSON} format only. The required \textbf{JSON} format is as follows:

\{ \\
 \hspace*{0.5em} "Ni": "property\_combination\_name", \\
 \hspace*{0.5em} "Zi": "A cohesive English sentence describing the property extraction pipeline." \\
\}
\end{prompt}
\vspace{0.3em}

\section{Dataset Examples}\label{app:dataset_examples}
In this section, we provide representative questions from \OURDATA{}, illustrating one example for each of the four difficulty levels.

\vspace{0.3em}
\begin{greenprompt}{Level-1}
\textbf{Question:}

For the molecule CCCCCCCCCCCCCCCCCC/C=C\textbackslash CCCCCCCC(=O)O, what is its molecular weight, and what is its Kappa1 topological descriptor?

\bigskip
\textbf{Parameters:}

CCCCCCCCCCCCCCCCCC/C=C\textbackslash CCCCC\\CCC(=O)O

\bigskip
\textbf{Standard\_Toolpath:}

 \hspace*{0.5em} ["SMILESToWeight", "GetKappa1"] \\

\end{greenprompt}
\vspace{0.3em}

\begin{greenprompt}{Level-2}
\textbf{Question:}

First convert the molecule names: Propionamide, Methane, Ethane, and Butane, to their respective SMILES representations. True or False: among these four molecules, the compound that contains at least one amide bond also has a larger number of heavy atoms than any of the other three.

\bigskip
\textbf{Parameters:}

Propionamide, Methane, Ethane, Butane

\bigskip
\textbf{Standard\_Toolpath:}

 \hspace*{0.5em} ["NameToSMILES", "GetAmideBondsNum", "GetHeavyAtomsNum"] \\

\end{greenprompt}
\vspace{0.3em}

\begin{greenprompt}{Level-3}
\textbf{Question:}

First convert the following SLN strings to SMILES: B-H-B3, CH3CH2C[*]HCH2Br, O[1]CH2CH(@1)CH3, C[1][H2]CH2CH2\\CH2CH2@1, CH3CH3, and C[1]:C:CH:\\CH:CH:CH:@1. Among the converted molecules, identify the one that simultaneously satisfies all of these conditions: (1) it is achiral, meaning its stereo code indicates no chirality; (2) it contains no ring system; (3) its Chi3v value is nonzero, indicating an acyclic branching pattern; and (4) it has the highest asphericity among those meeting conditions (2)--(4), indicating it is the least spherical.

\bigskip
\textbf{Options:}

A. B-H-B3

B. CH3CH2C[*]HCH2Br

C. O[1]CH2CH(@1)CH3

D. C[1][H2]CH2CH2CH2CH2@1

E. CH3CH3

F. C[1]:C:CH:CH:CH:CH:@1

\bigskip
\textbf{Parameters:}

B-H-B3, CH3CH2C[*]HCH2Br, O[1]CH2\\CH(@1)CH3, C[1][H2]CH2CH2CH2CH2\\@1, CH3CH3, C[1]:C:CH:CH:CH:CH:@1

\bigskip
\textbf{Standard\_Toolpath:}

 \hspace*{0.5em} ["SlnToSmiles", "GetStereoCodeFrom-\\Smiles", "GetRingsNum", "GetChi3v", "GetAsphericity"] 

\end{greenprompt}
\vspace{0.3em}

\begin{greenprompt}{Level-4}
\textbf{Question:}

A synthetic biology team is developing a compact DNA chassis for stable mammalian expression and wants to understand whether a newly generated sequence is a viable regulatory element rather than a problematic genomic insert. The sequence should be examined from sequence composition through structural plausibility and expression-related properties so that the team can decide whether it is worth advancing to experimental validation.

Generate one random DNA sequence. Predict whether it contains CpG islands that could trigger epigenetic silencing, and identify restriction enzyme recognition sites that could compromise cloning stability. Calculate its molecular weight in both double-stranded linear and single-stranded circular forms for construct planning. Identify the longest open reading frame, translate it into an amino-acid sequence, convert the resulting protein into a 3D structure, compute its hydrophobicity profile, and predict its overall solubility. Based on this integrated characterization, is the generated sequence more suitable as a benign engineered element or as a candidate likely to cause manufacturing or expression problems, and which features most strongly support that conclusion?

\bigskip
\textbf{Parameters:}

None

\bigskip
\textbf{Standard\_Toolpath:}

\hspace*{0.5em} ["RandomDNAGeneration", "CpGIslandPrediction", "SummaryEnzymeCleavageSites", "CalculateDoubleLinearDNAMolecularWeight", "CalculateSingleCircleDNAMolecularWeight", "FindORF", "TranslateDNAtoAminoAcidSequence", "SequenceToPdb", "ComputeProtScale", "ProteinSolubilityPredictor"] \\

\end{greenprompt}
\vspace{0.3em}



\end{document}